\documentclass[journal]{IEEEtran}
\usepackage{graphicx}
\usepackage{caption}
\usepackage{algorithmic}
\usepackage{algorithm}
\usepackage{comment}
\usepackage{bm}
\usepackage{subfigure}
\usepackage{color}
\usepackage{colortbl}
\definecolor{mygray}{gray}{0.9}
\usepackage[colorlinks,linkcolor=red]{hyperref}
\usepackage{multirow}
\usepackage{makecell}
\usepackage{booktabs}
\usepackage{amsmath}
\newtheorem{definition}{Definition}
\newtheorem{proof}{Proof}

\usepackage{array}
\usepackage{gensymb}
\usepackage{multicol}
\usepackage{subfigure}
\usepackage{amssymb}
\usepackage{mathrsfs}
\usepackage{color}
\usepackage{rotating}
\usepackage{threeparttable}
\usepackage{amssymb}
\usepackage{bigstrut}
\usepackage{verbatim}
\usepackage{bbding}
\usepackage[capitalize]{cleveref}
\usepackage[table]{xcolor}
\definecolor{deepred}{rgb}{0.698,0.133,0.133}
\usepackage{cite}
\crefname{section}{Sec.}{Secs.}
\Crefname{section}{Section}{Sections}
\Crefname{table}{Table}{Tables}
\crefname{table}{Tab.}{Tabs.}
\newcommand{\tabincell}[2]{\begin{tabular}{@{}#1@{}}#2\end{tabular}}
\newenvironment{myitemize}{\begin{list}{$\bullet$}
		{\setlength{\topsep}{1mm}
			\setlength{\itemsep}{0.25mm}
			\setlength{\parsep}{0.25mm}
			\setlength{\itemindent}{0mm}
			\setlength{\partopsep}{0mm}
			\setlength{\labelwidth}{15mm}
			\setlength{\leftmargin}{4mm}}}{\end{list}}

\ifCLASSINFOpdf
\else
\fi
\begin{document}
%
\title{Self-paced Weight Consolidation for\\Continual Learning}
%
%
%

\author{Wei~Cong,
	Yang~Cong,~\IEEEmembership{Senior~Member,~IEEE,}
	Gan~Sun,~\IEEEmembership{Member,~IEEE,}
	Yuyang~Liu, 
	Jiahua~Dong

\thanks{*The corresponding author is Prof. \textit{Yang Cong}. This work is supported by National Natural Science Foundation of China (Grant No. 62127807, 62003336) and National Postdoctoral Innovative Talents Support Program (BX20200353).}
\thanks{Wei Cong, Yuyang Liu and Jiahua Dong are with the State Key Laboratory of Robotics, Shenyang Institute of Automation, Chinese Academy of Sciences, Shenyang 110016, China, also with the Institutes for Robotics and Intelligent Manufacturing, Chinese Academy of Sciences, Shenyang 110169, China, and also with the University of Chinese Academy of Sciences, Beijing 100049, China (email: congwei@sia.cn; liuyuyang@sia.cn; dongjiahua@sia.cn).}
\thanks{Yang Cong and Gan Sun are with the State Key Laboratory of Robotics, Shenyang Institute of Automation, Chinese Academy of Sciences, Shenyang 110016, China, and also with the Institutes for Robotics and Intelligent Manufacturing, Chinese Academy of Sciences, Shenyang 110169, China (email: congyang81@gmail.com; sungan1412@gmail.com).}}
\maketitle

\begin{abstract}
Continual learning algorithms which keep the parameters of new tasks close to that of previous tasks, are popular in preventing catastrophic forgetting in sequential task learning settings. However, 1) the performance for the new continual learner will be degraded without distinguishing the contributions of previously learned tasks; 2) the computational cost will be greatly increased with the number of tasks, since most existing algorithms need to regularize all previous tasks when learning new tasks. To address the above challenges, we propose a \textbf{\underline{s}}elf-\textbf{\underline{p}}aced \textbf{\underline{W}}eight \textbf{\underline{C}}onsolidation (\textbf{spWC}) framework to attain robust continual learning via evaluating the discriminative contributions of previous tasks. To be specific, we develop a self-paced regularization to reflect the priorities of past tasks via measuring difficulty based on key performance indicator (\emph{i.e.,} accuracy). When encountering a new task, all previous tasks are sorted from ``difficult'' to ``easy'' based on the priorities. Then the parameters of the new continual learner will be learned via selectively maintaining the knowledge amongst more difficult past tasks, which could well overcome catastrophic forgetting with less computational cost. We adopt an alternative convex search to iteratively update the model parameters and priority weights in the bi-convex formulation. The proposed spWC framework is plug-and-play, which is applicable to most continual learning algorithms (\emph{e.g.,} EWC, MAS and RCIL) in different directions (\emph{e.g.,} classification and segmentation). Experimental results on several public benchmark datasets demonstrate that our proposed framework can effectively improve performance when compared with other popular continual learning algorithms.
\end{abstract}


%
\IEEEpeerreviewmaketitle

\section{Introduction}
%
%
%
%
\IEEEPARstart{C}{ontinual} learning~\cite{EWC} which focuses on overcoming catastrophic forgetting of model performance, has attracted widespread interests in the field of machine learning~\cite{pivot, obfcl} and computer vision, such as image recognition~\cite{tcsvt2}, object detection~\cite{tcsvt1, tcsvt4, CDT}, and semantic segmentation~\cite{PLOP,SDR,RCIL}. Inspired by human learning, continual learning studies how to learn from an infinite stream of data, with its two main goals as: 1) extend acquired knowledge gradually and use it for future learning~\cite{delange2021continual} and 2) reduce inference with previous tasks to prevent catastrophic forgetting~\cite{kemker2017fearnet}. For instance, a visual robot with continual learning ability should recognize the new coming \textit{table} well without forgetting the learned \textit{cat} and \textit{monitor}.
To realize continuous recognition of robots without catastrophic forgetting, several continual learning algorithms have been proposed, \emph{e.g.,} regularization-based algorithms~\cite{EWC, rannen2017encoder} penalize the importance parameters of previous tasks, replay-based algorithms~\cite{icarl, pgmr} take advantage of old samples in the current learning process, dynamic architecture-based algorithms~\cite{mallya2018packnet, aljundi2017expert} assign parameters for previous tasks dynamically, and latent knowledge-based algorithms \cite{ruvolo2013ella, 9524506, sun2020lifelong, sun2021and,tcsvt5} preserve a latent knowledge library of previous tasks, etc.

\begin{figure}[t]
	\centering
	\includegraphics[width=0.975\columnwidth]{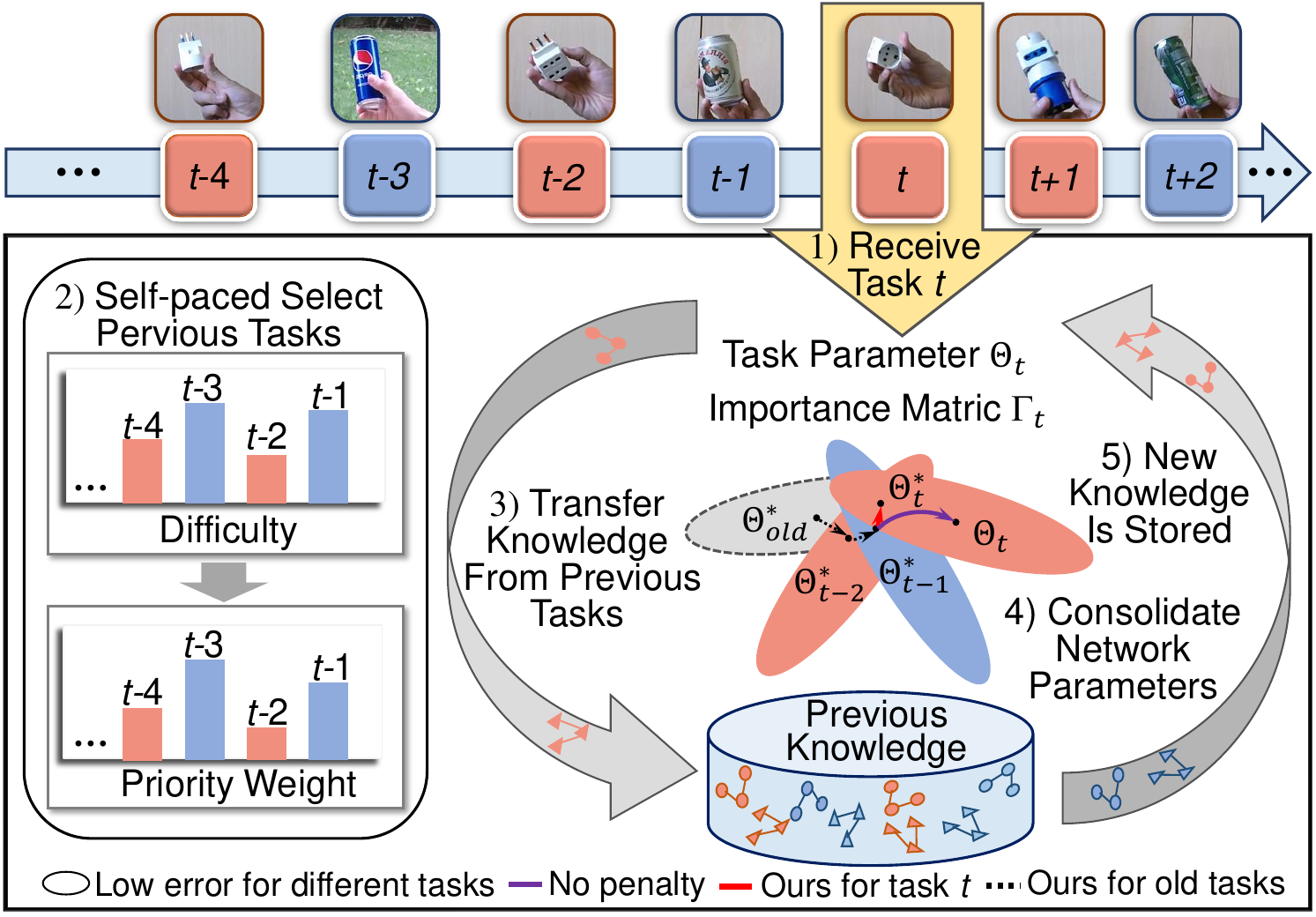} 
	\caption{Illustration of our self-paced Weight Consolidation framework for continual learning, where ${\Theta}_t$ denotes model parameters for task $t$. The same colors are similar in difficulty degree when learning the corresponding tasks, and vice versa.}
	\label{fig:motivation}
\end{figure}

To avoid storing any training data from past learned tasks, a number of regularization-based algorithms (\emph{e.g.,} EWC~\cite{EWC}, MAS~\cite{MAS} and RCIL~\cite{RCIL}) are developed to remember previous tasks by slowing down updating important knowledge for those tasks. More specifically, when learning a new task, EWC tries to preserve the model parameters close to the learned parameters amongst all previous tasks, which also suggests designing multiple penalties around the learning parameters. Inspired by how to gradually learn new knowledge by humans, some unreasonable aspects of regularization-based algorithms are revealed. 
For instance, \emph{when a programmer learns language \textit{C++}, he/she should spend the same time reviewing previously learned \textit{C} and \textit{python} to prevent catastrophic forgetting. Obviously, such review strategy is unreasonable since language \textit{C} is easier to remember than \textit{python} when learning \textit{C++}.} 
Based on the analysis above, we in this paper address the main challenges of the well-known regularization-based algorithms: 1) treating previous tasks equally affects the learning of new tasks, since the programmer should consolidate the language \textit{python} rather than \textit{C} when learning the language \textit{C++}; 
2) regularizing knowledge of all previous tasks to compute the regularization term increases the computational cost, due to the fact that it is unnecessary to consolidate the language \textit{C} for the programmer when learning \textit{C++}.

To solve above mentioned challenges, this paper establishes a new weight consolidation framework by integrating self-paced learning~\cite{kumar2010self} into continual learning, and taking into consideration the difficulties of previous tasks during learning new tasks. The concept of self-paced learning develops from curriculum learning~\cite{bengio2009curriculum} and originates from human education, including \emph{what to study, how to study, when to study,} and \emph{how long to study}~\cite{tullis2011effectiveness}. For the self-paced learning, its core idea is to learn simple samples before learning complex samples. In other words, self-paced learning can select a part of samples for training and give them different variable weights, which has been successfully applied to the fields of action detection~\cite{jiang2014self}, domain adaption~\cite{ge2020self} and feature selection~\cite{zheng2020unsupervised}. However, most current self-paced learning algorithms are implemented at the sample level, and its applications at the task level, \emph{i.e.,} control the order of learning tasks, is still scarce, especially in continual learning fields.

Motivated by the analysis above, as shown in ~\cref{fig:motivation}, we here propose a continual learning framework called self-paced Weight Consolidation, which explores to selectively consolidating the knowledge among past tasks to overcome catastrophic forgetting and simultaneously reducing the computational burden for the new task. The basic assumption for the framework is that the transferable knowledge from previous tasks makes discriminative contributions for the new continual learner while learning a new task. More specifically, we measure the difficulties of previous tasks based on the key performance indicator (\emph{i.e.,} accuracy). Then the priority weight for each past task can be reflected via a self-paced learning regularization. As a new task comes, the continual task learner can learn the new task by selectively backtracking ``difficult" past tasks and neglecting ``easy" ones according to priority weights. In addition, it is intuitively clear that the computational burden could be well mitigated with less penalties in comparison with traditional continual learning algorithms. To the end, experiments against other popular continual learning algorithms (\emph{e.g.,} EWC~\cite{EWC}, MAS~\cite{MAS}, RCIL~\cite{RCIL}) on several datasets strongly support the proposed framework. Moreover, the proposed spWC framework can be extended to other regularization-based continual learning algorithms. The contributions are summarized as follows:

\begin{myitemize}
	\item We propose a self-paced Weight Consolidation (spWC) framework, which could obtain a robust continual learner via selectively consolidating the knowledge amongst previously learned tasks with larger difficulties. 
	\item A self-paced regularization is designed in our spWC framework, which could assign a priority weight to each previous task based on the key performance indicator (\emph{i.e.,} accuracy). It further identifies ``difficult" or ``easy" to the new task.
	\item Some theoretical studies are derived to guarantee the rationality of the proposed spWC framework. Extensive experiments on several datasets show the superiority of our proposed spWC framework in terms of preventing catastrophic forgetting and mitigating computational cost.
	
\end{myitemize}

The rest of this paper is organized as follows. In ~\cref{related_work}, we review the related work on continual learning and self-paced learning. In ~\cref{self-paced}, we introduce the proposed spWC framework. In ~\cref{experiments}, we compare our proposed framework with other popular algorithms. Lastly, we conclude this paper and discuss the limitations in ~\cref{conclusion}.

\section{Related Work}
\label{related_work}

\subsection{Continual Learning}
Continual learning studies the problem of efficiently learning continuous tasks based on acquired knowledge without degrading performance on the previous tasks~\cite{chen2018lifelong}. According to the storage and utilization of task-specific information in the sequential learning process, we divide the recent continual learning algorithms into four categories: replay-based strategy, latent knowledge-based strategy, dynamic architecture-based strategy, and regularization-based strategy. The replay-based algorithms~\cite{icarl, lopez2017gradient, DER-Verse, pgmr,tcsvt3} learn a new task to alleviate forgetting by merging the stored or generated old samples into the current training process. Some latent knowledge-based algorithms~\cite{ruvolo2013ella, 9524506,Lifelong-Visual-Tactile,tcsvt5} are proposed to improve efficiency by preserving the learning experience of previous tasks in a latent knowledge library.~\cite{Jaehong2018, mallya2018packnet} dynamically assign parameters for previous tasks to guarantee the stability of performance. The regularization-based strategy~\cite{ILT, PLOP, RCIL, UCD,FCIL,FISS} is first developed by Elastic Weight Consolidation (EWC)~\cite{EWC}, which estimates the special distribution on the model parameters as the prior when processing continuous tasks. 
Meory aware synapses~\cite{MAS} redefines the parameter importance measure to an unsupervised setting. RCIL~\cite{RCIL} designs a pooled knowledge cube distillation strategy and a representation compensation module which decouples network branches into one frozen and one trainable.

Among the mentioned algorithms, we can notice that the replay-based strategy occupies a relatively high memory with the risk of leaking data privacy via reserving original data information~\cite{delange2021continual}. Besides, the latent knowledge-based strategy lacks a lot of theoretical basis in the construction of shared knowledge bases. Moreover, the dynamic architecture-based strategy continuously improves performance at the expense of memory consumption. While the regularization-based strategy pays attention to the correlation of the internal parameters or abstract meta-data to overcome catastrophic forgetting in continuous learning, which is closer to the synaptic consolidation process of the human brain. Undeniably, the regularization-based algorithms also exist problems, \emph{i.e.,} 1) the performance for the new continual learner will be degraded since previously learned tasks make equal contribution to the new task; 2) the computational cost will be greatly increased with the number of tasks by regularizing knowledge of all previous tasks when learning sequential tasks. We find that the core idea of self-paced learning is to learn ``easy'' knowledge before learning ``difficult'' knowledge. By setting an appropriate threshold to distinguish between ``easy'' and ``difficult'' tasks, self-paced learning can discriminatively treat previous tasks and reduce computational cost by ignoring ``easy'' tasks. Therefore, it is applicable to solve the above problems well by applying self-paced learning at the task level of continual learning.

\subsection{Self-paced Learning}
Bengio \emph{et al.} propose a new learning paradigm called curriculum learning~\cite{bengio2009curriculum}, which can gradually learn from simple and universal to difficult and complex samples. However, curriculum learning~\cite{PMLC} cannot make corresponding adjustments based on the feedback of the learners, since the curriculum is predetermined as a prior. To alleviate this deficiency,~\cite{kumar2010self, scg} design an iterative approach,\textit{ i.e.,} self-paced learning, which simultaneously selects easy samples and updates the parameters at each iteration. It differs from curriculum learning because the self-paced learner without prior knowledge can control what and when they learn. In order to integrate the diverse information of samples,~\cite{jiang2014self} proposes the algorithm of self-paced learning with diversity, which uses universal regularization to mine the preferences of easy and diverse samples. Many studies~\cite{li2017self, liChang2017self} extend self-paced learning by introducing group-wise weights to improve the performance on multiple data groups. For instance,~\cite{li2016multi} reformulates the self-paced learning problem as a multi-objective issue, which can obtain a set of solutions with different stopping criteria in a single run even under bad initialization. Furthermore,~\cite{jiang2015self} proposes a self-paced curriculum learning algorithm, which takes both the prior knowledge and the learning progress into account. {~\cite{lei_zhang_1, lei_zhang_2, lei_zhang_3} apply the idea of self-paced learning into different computer vision directions without special design for the self-paced regularization, which is meaningful but not applicable to our continual learning scenario.

Currently, most self-paced learning algorithms are implemented at the sample level. When faced with the situation of continual learning, how to effectively leverage self-paced learning at the task level is still a thought-provoking question waiting to be explored.

\section{Self-paced Weight Consolidation Framework}
\label{self-paced}
We first revisit the most popular continual classification algorithms, \emph{i.e.}, EWC~\cite{EWC} and MAS~\cite{MAS}, followed by combining them with our proposed spWC framework. Then we analyze the optimization and computational complexity. To further validate the generalization of our spWC, we lastly apply it to continual semantic segmentation scenario, \emph{i.e.,} RCIL~\cite{RCIL}.

\vspace{-1mm}
\subsection{Revisit EWC and MAS Algorithms}
A continual learning model faces a set of $m$ supervised learning tasks: $\mathcal{Z}_1,...,\mathcal{Z}_t,...,\mathcal{Z}_m$, where each individual task only contains training samples of one dataset. Specifically, the task $\mathcal{Z}_t=\{X_t , Y_t\}$ has $n_t$ training instances $X_t \in \mathbb {R}^{d \times n_t}$ with the corresponding labels $Y_t \in \mathbb {N}^{n_t}$. $d$ denotes the dimension of the training instance. Following the basic assumption of continual learning~\cite{parisi2019continual}, 1) the continual learning model does not know the total number, distribution and order of tasks; 2) the continual learning model can access $\mathcal{Z}_t$ only during the training phase of task $t$. Therefore, the goal of continual learning is to minimize the loss of the current task $\mathcal{Z}_m$ given only data $(X_m , Y_m)$ without accessing training samples of previous tasks $\mathcal{Z}_1,...,\mathcal{Z}_t,...,\mathcal{Z}_{m-1}$, which is formulated as:
\begin{equation}
	\mathcal{L}_m({\Theta}_m)=\mathcal{D}_{ce}(P_m(X_m; {\Theta}_{m}), Y_m),
	\label{eq_continual_learning}
\end{equation}
where $\mathcal{D}_{ce}(\cdot)$ represents the cross-entropy loss function, $P_m$ is the model output probability under the current task $\mathcal{Z}_m$ with the set of model parameters ${\Theta}_m\in\mathbb {R}^{N}$.

As the founder of continual learning algorithms, EWC develops a quadratic penalty for the discrepancy between the parameters of the new and the old tasks. Specifically, EWC adopts the Fisher information matrix to measure the importance of each parameter of the previous tasks. EWC can achieve the goal to learn the new task $\mathcal{Z}_m$ without forgetting previous tasks $\mathcal{Z}_{1},...,\mathcal{Z}_{m-1}$ by minimizing the loss function $\mathcal{L}({\Theta}_m)$, which is formulated as the following:
\begin{equation}
	\mathcal{L}({\Theta}_m)= \mathcal{L}_m({\Theta}_m)+\sum_{t=1}^{m-1}\sum_{i}\frac{\lambda}{2}\Gamma_{t,i}(\Theta_{m,i}-\Theta_{t,i}^*)^2,
	\label{eq_ewc}
\end{equation}
where $\lambda$ is the regularization parameter, $i$ labels each parameter of the classification model. Moreover, ${\Theta}_{m,i}$ is the parameter $i$ for the new task $\mathcal{Z}_m$ and $\Theta_{t,i}^*$ is the optimal parameter $i$ of any old task $\mathcal{Z}_t$. $\Gamma_{t,i}$ represents the importance of parameter $i$ to any old task $\mathcal{Z}_t$, which adopts FIsher information matrix as the measurement. Therefore, EWC can selectively slow down the update of parameters important for previous tasks to avoid catastrophic forgetting.

Similarly, the objective function of MAS is as follows:
\begin{equation}
	\mathcal{L}({\Theta}_m)= \mathcal{L}_m({\Theta}_m)+ \sum_{i,j}\gamma\Omega_{m-1,ij}(\Theta_{m,ij}-\Theta_{m-1,ij}^*)^2,
	\label{eq_mas}
\end{equation}
where $\Theta_{m,ij}$ of the current model denotes the parameters of the connections between pairs of neurons $n_i$ and $n_j$ in two consecutive layers. $\Theta_{m-1,ij}^*$ represents the optimal parameters of the connections between pairs of neurons $n_i$ and $n_j$ in two consecutive layers for the previous model. $\Omega_{m-1,ij}$ denotes the importance of the parameter $\Theta_{m-1,ij}^*$ by accumulating gradients over the given data points, which is updated after training a new task. $\gamma$ is the regularization parameter.

However, the continual learning algorithms (\emph{e.g.,} EWC) expose the following shortcomings when learning a new task. 1) Since each learned task $\mathcal{Z}_t$  has a different data distribution, the differences  between each previous task $\mathcal{Z}_t$ and the new task $\mathcal{Z}_m$ are treated consistently in Eq. \eqref{eq_ewc}, which could degrade the model performance. 2) In addition, the quadratic penalty term regularizes knowledge of all previous tasks during the training phase of the new task $\mathcal{Z}_m$, which makes the computational cost greatly increase as the number of tasks. In the next subsection, we consider the self-paced learning which can select the knowledge amongst previously ``difficult'' tasks to consolidate based on the difficulty of past tasks.

\subsection{Our Proposed spWC Framework}
This subsection introduces our proposed spWC framework in the perspective of different contribution of the previous tasks to the new continual learner, \emph{i.e.,} the knowledge of previously learned tasks is not equal contribution to the new learner in the continual learning process.

When facing a new task, the continual learner should capture the difficulty of past tasks based on key performance indicator (\emph{i.e.,} accuracy). Then all previous tasks are sorted from ``difficult'' to ``easy'' based on the priorities. The parameters for the new continual learner will be learned via selectively maintaining the knowledge among previously ``difficult'' tasks, \emph{i.e.,} assigning discriminative priority weights of the previously selected tasks to the new learner. Specifically, we achieve this challenge by incorporating a self-paced regularization into the learning objective of continual learning. The objective function of applying our proposed spWC framework to the representative EWC algorithm can be formulated as the following:
\begin{equation}
	\begin{split}
		\mathcal{L}({\Theta}_m)= &\mathcal{L}_m({\Theta}_m)+\sum_{t=1}^{m-1}\sum_{i}\frac{\lambda}{2}v_t\Gamma_{t,i}(\Theta_{m,i}-\Theta_{t,i}^*)^2\\
		&+ \mu_mf({v};m).
		\label{eq_spwc}
	\end{split}
\end{equation}
The objective function of applying our proposed spWC framework to the popular MAS algorithm can be formulated as:
 \begin{equation}
	\begin{split}
		\mathcal{L}({\Theta}_m)= &\mathcal{L}_m({\Theta}_m)+\sum_{t=1}^{m-1}\sum_{i,j}\gamma v_t \Omega_{t,ij}(\Theta_{m,ij}-\Theta_{t,ij}^*)^2\\
		&+ \mu_mf({v};m),
		\label{eq_spwc_mas}
	\end{split}
    \end{equation}
where $f({v};m)$ represents the self-paced regularization with respect to ${v}$ and $m$. $v_t \in [0,1]$ denotes the priority weight of task $t$ and ${v}=[v_1,...,v_{m-1}]$ is the vector of priority weights. The age parameter $\mu_m$ is the age to control the learning pace.

Next we introduce our design of the \textit{Self-paced Regularization} $f({v};m)$ and derive the \textit{Priority Weight} $v_t$. Furthermore, we will provide the \textit{Theoretical Proof} of our proposed self-paced regularization. Note that we use the representative EWC via applying our spWC framework to make following analysis.

\subsubsection{Self-paced Regularization} The initial self-paced regularization~\cite{jiang2014easy} leads to a binary weighting on the samples, assigning $1$ to ``easy'' samples and $0$ to ``difficult'' samples. Since no two ``easy'' (or ``difficult'') samples are likely to be strictly equally significant and learnable, this form of binary weights tends to lose flexibility. Compared to the binary regularization, the solutions of various soft regularizations assign soft weights to reflect samples importance in finer granularity, which helps soft regularizations achieve better performance in various applications. However, all these techniques: 1) adopt self-paced learning to distinguish difficulty between samples, \emph{i.e.,} implement self-paced learning at the sample level; 2) assign high weights to the ``easy'' samples but low weights to the ``difficult'' samples, that is, more energy is spent on learning ``easy'' knowledge, which is not in line with human learning habits. To address this issue, we design a self-paced regularization which 1) employs self-paced learning to differentiate varying degrees of difficulty between tasks, \emph{i.e.,} carry out self-paced learning at the task level; 2) reasonably assigns weights (\emph{i.e.,} assign low weights to the ``easy'' samples and high weights to the ``difficult'' samples) by encoding latent relevance between tasks to effectively learn the new task while overcoming catastrophic forgetting. The proposed self-paced regularization $f({v};m)$ is formulated as the following:
\begin{equation}
	f({v};m)= \frac{1}{3}||{v}||_2^{3}-\sum_{t=1}^{m-1}v_t.
	\label{eq_regularization}
\end{equation}
Our designed self-paced regularization allows the generated soft weights to have a reasonable degree of discrimination (not too large or too small). Moreover, it bridges the gap in designing an appropriate self-paced regularization term for continuous learning scenario.
Next, we derive priority weights of past tasks based on difficulty degree according to Eq. \eqref{eq_regularization}.

{\subsubsection{Priority Weight} To simplify the learning objective of spWC (\emph{i.e.,} Eq. \eqref{eq_spwc}), let $l_t$ represent the weighted distance between the model parameters of the learned task $t$ and the model parameters of the new task:
\begin{equation}
	l_t=\sum_{i}\frac{\lambda}{2}\Gamma_{t,i}(\Theta_{m,i}-\Theta_{t,i}^*)^2.
	\label{l_t}
\end{equation} 
It can be seen from Eq. \eqref{l_t}, $l_t$ indirectly reflects the difficulty of not forgetting task $t$ when encountering a new task. By replacing the corresponding part in Eq. \eqref{eq_spwc} with Eq. \eqref{l_t}, we can obtain the optimal priority weight by selecting tasks with greatest distance between the FIsher information of the new task and the previous tasks, which is shown as follows:
\begin{equation}
	v_t^*(l_t;\mu_m)=\arg \mathop{\max}_{v_t}\ v_tl_t + \mu_m(\frac{1}{3}||v_t||_2^{3}-v_t).
    \label{eq_simplify}
\end{equation}

For the self-paced learning, its core idea is to learn simple samples before learning complex samples. In addition, it selects a part of samples for training and gives larger weights to easier samples. On the contrary, our spWC focuses more on the parameters of difficult tasks and assign larger weights to more difficult tasks. In order to obtain $v_t^*$ more intuitively and accurately, we adopt the difficulty $\eta_t$ of task $\mathcal{Z}_t$ to represent $l_t$ in \cref{eq_simplify} when encountering a new task. In particular, the difficulty $\eta_t$ of task $\mathcal{Z}_t$ can be calculated according to the key performance indicator $\psi_t$, which is formulated as follows:
\begin{equation}
	\eta_t=-\psi_t\log(1-\psi_t).
	\label{eq_D}
\end{equation}
When encountering the task $\mathcal{Z}_{m}$, we only need the current model parameters inherited from the task $\mathcal{Z}_{m-1}$ to calculate the test accuracy of the past task $t$ as the key performance indicator $\psi_t$. spWC only needs to store a list $\psi_t$ with several elements whose memory space occupation can be ignored. Compared with $l_t$, $\eta_t$ can directly reflect the performance of the model on the previous task $t$, and the design of $\eta_t$ makes the priority weight more in line with human learning habits, that is, gives larger weight to the difficult task and vice versa.
By replacing $l_t$ with $\eta_t$, Eq. \eqref{eq_simplify} can be rewritten as:
\begin{equation}
	v_t^*(\eta_t;\mu_m)=\arg \mathop{\min}_{v_t}\ v_t\eta_t + \mu_m(\frac{1}{3}||v_t||_2^{3}-v_t).
	\label{eq_diff}
\end{equation}
Eq. \eqref{eq_diff} is a convex function of $v_t$ in [0,1] and thus the global minimum can be obtained as the following:
\begin{equation}
		\eta_t+ \mu_m(v_t^{2}-1)=0.
	\label{eq_jie}
\end{equation}
By solving Eq. \eqref{eq_jie}, the closed-form optimal solution for $v_t$ can be obtained as follows:	

\begin{equation}
	v_t^*= 
	\left\{
	\begin{array}{lr}
		(1-\frac{\eta_t}{\mu_m}) ^{\frac{1}{2}} & \eta_t<\mu_m\\
		0 & \eta_t\ge\mu_m,
	\end{array}
 \right.
 \label{eq_weight}
\end{equation}}

\subsubsection{Theoretical Proof} We will prove the theoretical rationality of our self-paced regularization below. Basically, an eligible self-paced regularization is desired to match the requirements illustrated in Definition~\ref{definition}~\cite{jiang2014easy}.
		
\begin{definition}
	Suppose $v,l$ and $\mu$ are the sample weight, sample loss, and age parameter, respectively. $f$ is called a self-paced regularization, if the following statements hold:
	
	1) $f$ is convex with respect to $v\in [0,1]$;
	
	2) $v^*(l,\mu)$ is monotonically decreasing with respect to $l$, and $\lim_{l\rightarrow 0} v^*(l,\mu)=1$, $\lim_{l\rightarrow \infty} v^*(l,\mu)=0$;
	
	3) $v^*(l,\mu)$ is monotonically increasing with respect to $\mu$, and $\lim_{\mu\rightarrow 0} v^*(l,\mu)=0$, $\lim_{\mu\rightarrow \infty} v^*(l,\mu)\le 1$;
	where $v^*(l,\mu)=\arg\min_{v\in [0,1]} vl+f$.
	\label{definition}
\end{definition}

Some previous works~\cite{multi-task} have proved that Definition~\ref*{definition} is applicable to the task level. We will prove the rationality of our proposed self-paced regularization based on this definition, which is provided as follows:

\begin{proof} The first and second derivatives of Eq. \eqref{eq_regularization} for $v_t$ are expressed as the following:
\begin{equation}
	\left\{
	\begin{array}{lr}
		\frac{\partial f(v_t;m)}{\partial v_t}=v_t^{2}-1  \\
		
		\frac{\partial^2 f(v_t;m)}{\partial^2 v_t}=2v_t.
	\end{array}
	\right.
	\label{eq9}
\end{equation}
The second derivative of Eq. ~\eqref{eq_regularization} is $\frac{\partial^2 f({v}_t;m)}{\partial^2 {v_t}}\geqslant0$, so $f({v};m)$ is convex with respect to $v_t\in [0,1]$. Therefore, condition 1 of Definition~\ref*{definition} holds.
According to Eq. \eqref{eq_weight}, the following conclusions can be drawn: $\lim_{\eta_t\rightarrow 0} v_t^*(\eta_t;\mu_m)=1$; $\lim_{\eta_t\rightarrow \infty} v_t^*(\eta_t;\mu_m)=0$; $\lim_{\mu_m\rightarrow 0} v_t^*(\eta_t;\mu_m)=0$; $\lim_{\mu_m\rightarrow \infty} v_t^*(\eta_t;\mu_m)\le 1$. 
As a result, the conditions 2 and 3 of Definition~\ref*{definition} also hold.
\end{proof}

Consequently, our proposed self-paced regularization meets all the requirements of Definition~\ref*{definition}. Therefore, we get the conclusion that it is a rational self-paced regularization.

\begin{table*}[htb]
	
	\caption{List of used classification datasets: Permuted-MNIST, CIFAR-100 and Tiny Imagenet.}

	\scalebox{1.16}{
		\begin{tabular}{l|llllll}
			\toprule
			& Tasks & Classes/task & Train data/task & Valid data/task & Test data/task &\makecell[c]{Task selection}\\ \midrule
			Permuted-MNIST~\cite{EWC} & \makecell[c]{10}    & \makecell[c]{10}           & \makecell[c]{40000}           & \makecell[c]{10000}           & \makecell[c]{10000}          & \makecell[c]{all classes via permutation}\\ 
			CIFAR-100~\cite{krizhevsky2009learning}      & \makecell[c]{10}    & \makecell[c]{10}           & \makecell[c]{4000}            & \makecell[c]{1000}            & \makecell[c]{1000}           & \makecell[c]{random classes} \\ 
			Tiny-Imagenet~\cite{wu2017tiny}  & \makecell[c]{10}    & \makecell[c]{20}           & \makecell[c]{8000}            & \makecell[c]{2000}            & \makecell[c]{1000}           & \makecell[c]{random classes}\\ \bottomrule
			
		\end{tabular}
	}
	
	\label{dataset}
\end{table*}

\begin{algorithm}[t]
	\caption{Our spWC Framework to EWC}
	\label{alg:algorithm}
	\textbf{Input}: Sequential supervised tasks $\mathcal{Z}_1, \mathcal{Z}_2,...,\mathcal{Z}_m$, Regularization parameter $\lambda\ge 0$, Age parameter $\mu_m\ge0$;\\
	\textbf{Output}: Model parameter ${\Theta}^*$;
	\begin{algorithmic}[1] 
		\IF {isSupervisedTaskAvailable()}
		\IF {$m==1$}
		\STATE ${\Theta}_m\gets \arg\min\limits_{\Theta{_m}}\mathcal{L}({\Theta}_m;{\Theta}_{m-1},\mathcal{Z}_m)$;
		\STATE ${\Gamma_m} \gets$ getFIsherMatrix$({\Theta}_m)$;
		\ELSE
		
		\FOR {$t=1,...,m-1$}
		
		\STATE ClassificationNet $\gets$ BackboneNet$({\Theta}_m)$;
		\STATE $\psi_t \gets$ getAccuracy(ClassificationNet, $\mathcal{Z}_t)$;\\
		\STATE $\eta_t = -\psi_tlog(1-\psi_t)$ via Eq.~\eqref{eq_D};\\
		\IF {$\mu_m\textgreater \eta_t$}
		\STATE  ${v}_t^*= (1-\frac{\eta_t}{\mu_m}) ^{\frac{1}{2}}$ via Eq.~\eqref{eq_weight1};\\
		\ELSE
		\STATE  ${v}_t^*= 0$ via Eq.~\eqref{eq_weight1};\\
		\ENDIF
		\ENDFOR
		
		\STATE ${\Theta}_m\gets$ GradientOptimization$(\{{\Theta}_t, {\Gamma}_t, v_t^*\}_{t=1}^{m-1},\lambda)$ via Eq.~\eqref{eq_theta};\\
		\STATE ${\Gamma_m} \gets$ getFIsherMatrix$({\Theta}_m)$;
		\ENDIF
		\ENDIF
		\STATE \textbf{return} ${\Theta}^*={\Theta}_m$.\\
		
	\end{algorithmic}
    \label{algorithm_1}
\end{algorithm}

\subsection{Model Optimization}
To optimize the objective function in Eq.~\eqref{eq_spwc}, we use an alternative convex search~\cite{gorski2007biconvex} strategy, which is an iterative algorithm for biconvex optimization, and the variables are divided into two disjoint blocks. In each iteration, one block of variables is optimized while keeping the other block fixed. The optimization progress of our spWC framework to EWC algorithm is summarized in {\bf Algorithm~\ref{algorithm_1}}. After initializing the model parameters when $m=1$, the updating procedure can be presented as follows.   


{\bf Updating ${v}$ with the fixed ${\Theta}_m$:} For the new coming $m$-th task, we keep ${\Theta}_m$ fixed and update the priority weight ${v}$. In order to realize this goal, priority weight $v_t$ of each previous task is calculated based on the difficulty $\eta_t$. During the training phase of new task, we only update ${v}$ once and the global optimum of the priority weight ${v}^*=[v_1^*,...,v_{m-1}^*]$ can be calculated as follows:
\begin{equation}
	v_t^*= 
	\left\{
	\begin{array}{lr}
		(1-\frac{\eta_t}{\mu_m}) ^{\frac{1}{2}} & \eta_t<\mu_m\\
		0 & \eta_t\ge\mu_m,
	\end{array}
	\right.
	\label{eq_weight1}
\end{equation}

{\bf Updating  ${\Theta}_m$ with the fixed ${v}^*$:} For the new coming $m$-th task, we keep the global optimum of the priority weight ${v}^*=[v_1^*,...,v_{m-1}^*]$ fixed and update the model parameter ${\Theta}_m$. During the training phase of new task, the global optimum of ${\Theta}_m^*$ can be expressed as follows:
\begin{equation}
	{\Theta}_m^*=\arg \min\limits_{\Theta{_m}} \mathcal{L}_m({\Theta}_m)+\sum_{t=1}^{m-1}\sum_{i}\frac{\lambda}{2}v_t^*\Gamma_{t,i}(\Theta_{m,i}-\Theta_{t,i}^*)^2.
	\label{eq_theta}
\end{equation}
Through the alternative update and optimization of ${\Theta}_m$ and ${v}$, we can obtain the optimal performance of the proposed spWC.

\subsection{Computational Complexity}
When learning a new task, the computational complexity for our spWC framework to EWC involves two conditions: 1) When the number of tasks is $1$ (\emph{i.e.,} m=1), the main computational cost of spWC is to update $\Theta_m$ in Eq.~\eqref{definition}. More specifically, the cost of updating $\Theta_m$ is $O(\xi(N,n_m))$, where $\xi(\cdot)$ depends on the single-task learner (i.e., the base model), and $n_m$ denotes the number of training samples, $N$ denotes the number of model parameters. Therefore, when a new task is coming, the overall computational complexity of our proposed model is $O(\xi(N,n_m))$. 
2) When the number of tasks is greater than $1$ (\emph{i.e.,} $m>1$), the main computational cost of learning a new task involves two subproblems: one optimization problem lies in Eq~\eqref{eq_weight1}, the other one is in Eq~\eqref{eq_theta}. For the problem in Eq~\eqref{eq_weight1}, the cost of computing $\psi_t$ is $O((m-1)\zeta(N,e_m))$, where $\zeta(\cdot)$ represents the testing process of each task and $e_m$ denotes the number of testing samples. The testing process only contains one neural network forward propagation by using small number of testing samples (\emph{i.e.,} the number of testing samples is one quarter of the number of training samples), whose computational cost is trivial compared with the training process. Next, computing $\eta_t$ and $v_t^*$ cost $O(m-1)$. For the problem in Eq~\eqref{eq_theta}, the cost of updating $\Theta_m$ is $O(\xi(N,n_m)+kNn_m)$, where $O(kNn_m)$ represents the computational complexity of calculating the regularization term, and $k$ is the number of previous tasks selected by spWC and $k < m-1$. Since $O((m-1)\zeta(N,e_m )+(m-1))$ is negligible compared with $O(\xi(N,n_m )+kNn_m)$, the overall computational complexity of our proposed algorithm is $O(\xi(N,n_m )+kNn_m)$.

\subsection{Our spWC Framework to Continual Semantic Segmentation}
For continual semantic segmentation (CSS), the task $\mathcal{Z}_t=\{X_t , Y_t\}$ has $n_t$ input image $X_t$ with the corresponding segmentation ground-truth $Y_t$. When training on $\mathcal{Z}_t$, the training data of previous classes are not accessible. Besides, the ground-truth in $\mathcal{Z}_t$ only contains $C_t$ new classes, while the old and future classes are labeled as background. The goal of CSS is to predict all the classes $C_{0:m}$ seen over time.

As a popular algorithm in CSS, RCIL~\cite{RCIL} proposes RC module consisted of two dynamically evolved branches with one frozen and one trainable, which decouples the representation learning of both old and new knowledge. Besides, it designs a pooled cube knowledge distillation strategy on both spatial and channel dimensions to enhance the plasticity and stability of the model. The objective function of RCIL with pseudo labels can be formulated as:
\begin{equation}
	\mathcal{L}({\Theta}_m)=\mathcal{D}_{ce}(P_m(X_m; {\Theta}_{m}), \hat{Y}_m) + L_{skd} + L_{ckd},
	\label{eq_semantic_segmentation}
\end{equation}
where $L_{skd}$ and $L_{skd}$ denote the spatial and channel knowledge distillation loss, respectively. $\hat{Y}_m$ is the pseudo label of $X_m$.
Then the objective function of applying our spWC framework to RCIL algorithm can be formulated as follows:
 \begin{equation}
	\begin{split}
		\mathcal{L}({\Theta}_m) = & v\mathcal{D}_{ce}(P_m(X_m; {\Theta}_{m}), \hat{Y}_m) + L_{skd} + L_{ckd}\\
		&+ \mu_mf({v};m),
		\label{eq_spwc_rcil}
	\end{split}
    \end{equation}
where $f({v};m)$ represents the self-paced regularization with respect to priority weights ${v}$ and the current task $m$. The age parameter $\mu_m$ is the age to control the learning pace.
\section{Experiments}
\label{experiments}
This section evaluates our proposed spWC framework via several empirical comparisons on several benchmark datasets. We firstly describe the used competing algorithms, evaluation metrics and datasets. Extensive experimental results are then conducted, followed by some discussions and analysis.

\subsection{Experimental Setup}
\subsubsection{Comparison  Algorithms}
The experiment in this subsection evaluates our proposed spWC framework. All continual learning comparison algorithms in our experiments are as the following: \textbf{Finetune} is sequentially trained on the new coming task in the standard way, which can be seen as a lower bound. \textbf{JointTrain} is always trained using samples of all tasks so far, which can be seen as an upper bound. \textbf{EWC}~\cite{EWC} approximates the parameters update constraint of a posterior Gaussian distribution. \textbf{LWF}~\cite{LWF} transfers knowledge by distilling knowledge from a previous model. For continual classification: \textbf{SI}~\cite{zenke2017continual} estimates  importance weights online over the entire learning trajectory in the parameter space. \textbf{MAS}~\cite{MAS} suggests unsupervised importance estimation, allowing increased flexibility. \textbf{BLIP}~\cite{BLIP} estimates information gained on each parameter to determine which bits to be frozen. \textbf{EWC+spWC} applies spWC framework to EWC~\cite{EWC} algorithm which is explained in ~\cref{self-paced}. \textbf{MAS+spWC} applies spWC framework to MAS~\cite{MAS} algorithm. spMS only retains the way MAS calculates important weights and the rest are consistent with EWC. For continual semantic segmentation: \textbf{LWF-MC}~\cite{icarl} replays the closest old samples to each class's feature mean and distills the old model's prediction to that of the current model. Moreover, we conduct experiments on several state-of-the-art incremental semantic segmentation (ISS) algorithms. \textbf{ILT}~\cite{ILT} distills the previous model's intermediary features and output probabilities to the current model. \textbf{MiB}~\cite{MIB} converts the likelihood of the background to the probability sum of old or future classes for handling background shift. \textbf{PLOP}~\cite{PLOP} distills multi-scale features and labels the background based on entropy. \textbf{ALIFE}~\cite{alife} presents a feature replay scheme and an adaptive logit regularizer to balance model accuracy and efficiency. \textbf{RCIL}~\cite{RCIL} decomposes the representation learning into the old and new knowledge, and designs a pooled cube knowledge distillation strategy. \textbf{RCIL+spWC} applies our proposed spWC framework to RCIL~\cite{RCIL} algorithm.


\subsubsection{Evaluation Metrics}
The following metrics are used to demonstrate the model performance on sequential tasks during the test phase. For the continual classification problem: \textit{\textbf{Average Per-Task Accuracy (APA)}}~\cite{Jaehong2018,lee2019learning}: the average test accuracy of all previously seen tasks under the current one, \emph{i.e.,} the overall performance of the current model on all seen tasks. 
\textit{\textbf{Average Catastrophic Forgetting (ACF)}}: The discrepancy between the expectation of acquired knowledge of all tasks, \emph{i.e.,} the average accuracy after learning the first task and the average accuracy after training one or more additional tasks. \textit{ACF} can more intuitively present the continual learning ability of the entire network, regardless of its own performance on each task. \textit{\textbf{Parameters Size (PS)}} efficiency: The parameters size Prams($\Theta_t$) of each model is quantified by parameters at task $\mathcal{Z}_t$. Prams($\Theta_t$) should grow slowly relative to Prams($\Theta_1$). Formally, we design \textit{PS} as:
	\begin{equation}
		PS=\min(1,\frac{\sum_{t=1}^{m}\frac{{\rm Prams}(\Theta_1)}{{\rm Prams}(\Theta_t)}}{m}),
		\label{PS}
	\end{equation}
where $m$ is the total number of tasks. \textit{PS} demonstrates the classification efficiency and reflects the size of the computational memory. The larger the \textit{APA}, the smaller the \textit{ACF}, the larger the \textit{PS}, the better the performance of the classification model. For continual semantic segmentation: We use \textbf{mean Intersection over Union (mIoU)} to measure the segmentation performance. Three results are computed after the last task $m$: 1) mIoU for the initial classes $\mathcal{C}_0$, which reflects the competence to overcome catastrophic forgetting; 2) mIoU for the incremental classes $\mathcal{C}_{1:m}$ for measuring the capacity of adapting to new classes; 3) mIoU for all classes $\mathcal{C}_{0:m}$, which measures the overall continual semantic segmentation ability.

\begin{table*}[ht]\centering
	\caption{{ Comparison Of All Models on Permuted-MNIST, CIFAR-100 and Tiny Imagenet Datasets in terms of Average-\textit{Average Per-Task Accuracy (APA)} and Average-\textit{Average Catastrophic Forgetting (ACF)}, Where the Top Value Is Highlighted by \textbf{\color{purple}Red Bold} Font and The Runner-up by \textbf{\color{blue}Blue Bold} Font.}}
	\scalebox{1}{		
		\begin{tabular}{c|c|c|c|c|c|c}
			\toprule
			Datasets   & \multicolumn{2}{c|}{Permuted-MNIST~\cite{EWC}} & \multicolumn{2}{c|}{CIFAR-100~\cite{krizhevsky2009learning}}  & \multicolumn{2}{c}{Tiny Imagenet~\cite{wu2017tiny}} \\
			Metrics    & Average-\textit{APA} (\%)   & Average-\textit{ACF} (\%)  & Average-\textit{APA} (\%) & Average-\textit{ACF} (\%) & Average-\textit{APA} (\%)   & Average-\textit{ACF} (\%) \\ \midrule
			Finetune   & 85.4$\pm$1.1              & 12.3$\pm$1.0         & 29.4$\pm$3.3            & 37.2$\pm$2.6       & 29.6$\pm$2.5              & 24.3$\pm$2.1        \\
			JointTrain & 96.2$\pm$0.0              & $\enspace$0.3$\pm$0.0          & 63.0$\pm$0.1             & $\enspace$0.1$\pm$0.0        & 50.3$\pm$0.3              & $\enspace$1.2$\pm$0.1         \\
			EWC~\cite{EWC}        & 93.0$\pm$0.1              & $\enspace$3.9$\pm$0.2          & 51.3$\pm$1.9            & 13.1$\pm$2.0       & 42.9$\pm$2.2              & $\enspace$9.4$\pm$2.0         \\
			online-EWC~\cite{schwarz2018progress}        & 93.5$\pm$0.3              & $\enspace$3.3$\pm$0.3          & 50.5$\pm$2.5            & 14.0$\pm$2.6       & 42.3$\pm$2.3              & 10.1$\pm$2.8         \\
			LWF~\cite{LWF}        & 93.9$\pm$0.1              & $\enspace$2.9$\pm$0.1          & 54.4$\pm$1.0            & $\enspace$9.7$\pm$0.5        & 43.8$\pm$0.9              & $\enspace$8.4$\pm$1.1         \\
			SI~\cite{zenke2017continual}         & 93.6$\pm$0.3              & $\enspace$3.2$\pm$0.2          & 52.7$\pm$1.9            & 11.5$\pm$1.7       & 42.2$\pm$2.0              & 10.3$\pm$1.7        \\
			MAS~\cite{MAS}        & 91.1$\pm$0.2              & $\enspace$6.0$\pm$0.2          & 51.1$\pm$1.8            & 13.3$\pm$2.0       & 40.3$\pm$1.5              & 12.3$\pm$1.6        \\
			BLIP~\cite{BLIP}       & 93.8$\pm$0.1              & $\enspace$3.0$\pm$0.1          & 51.3$\pm$0.4            & 11.4$\pm$0.4       & 43.5$\pm$0.5              & $\enspace$8.7$\pm$0.5         \\ \midrule
			\textbf{EWC~\cite{EWC}+spWC (Ours)}       & \textbf{\color{purple}95.1$\pm$0.1}              & $\enspace$\textbf{\color{purple}1.6$\pm$0.1}          & \textbf{\color{blue}55.3$\pm$0.3}            & $\enspace$\textbf{\color{blue}8.7$\pm$0.2}        & \textbf{\color{purple}46.1$\pm$0.3}              & $\enspace$\textbf{\color{purple}5.9$\pm$0.4}         \\ 
			\textbf{MAS~\cite{MAS}+spWC (Ours)}      & \textbf{\color{blue}94.8$\pm$0.1}              & $\enspace$\textbf{\color{blue}1.9$\pm$0.1}          & \textbf{\color{purple}55.9$\pm$0.3}            & $\enspace$\textbf{\color{purple}8.1$\pm$0.4}        & \textbf{\color{blue}45.7$\pm$0.2}              & $\enspace$\textbf{\color{blue}6.4$\pm$0.3}      \\ \bottomrule
		\end{tabular}
	}
	\label{table_accuracy}
\end{table*}

\begin{figure*}[htb]
	\centering
	\includegraphics[width=2.04\columnwidth]{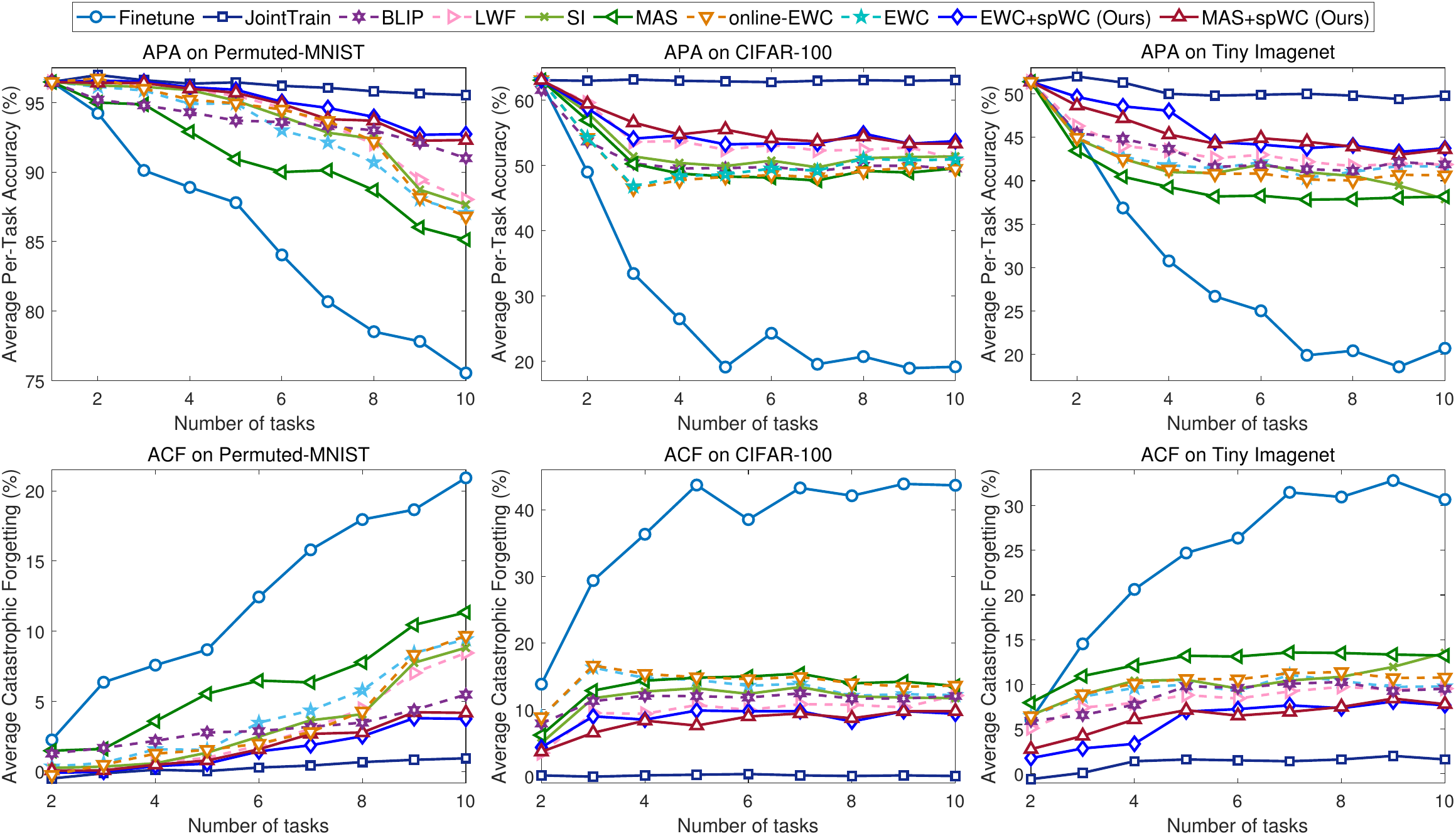} 
	\caption{{The results for state-of-the-art competing algorithms in terms of \textit{Average Per-Task Accuracy (APA)} (\textbf{Top}) and \textit{Average Catastrophic Forgetting (ACF)} (\textbf{Bottom}) on three public datasets, \emph{i.e.,} Permuted-MNIST (\textbf{Left}), CIFAR-100 (\textbf{Middle}) and Tiny Imagenet (\textbf{Right}), after training each task. Lines with different colors represent different comparison models.}}
	\label{he}
\end{figure*}

\subsubsection{Datasets}
We evaluate our proposed spWC framework on four publicly representative datasets. For continual classification: the main characteristics of three datasets in our experiments are summarized in ~\cref{dataset}. \textbf{Permuted-MNIST}~\cite{EWC} contains ten tasks, where each task is now a ten-way classification. It is the variant of MNIST~\cite{larochelle2007empirical} dataset which contains 10 different classes and each image is a grey scale image in size 28x28. To generate the permutated images, the original images are first zero-padded to 32x32 pixels. For each task, a random permutation is then generated and applied to these 1024 pixels. For each task, the dataset is consisted of 40000 training samples, 10000 valid samples and 10000 test samples. \textbf{CIFAR-100}~\cite{krizhevsky2009learning} is consisted of ten tasks, where each task is now a ten-way classification. The CIFAR-100 dataset, as a subset of the Tiny Images dataset~\cite{torralba200880}, is composed of 100 different classes. Specifically, each image in this dataset is an RGB image in size 32×32. For each task, the dataset is divided into 4000 training samples, 1000 valid samples and 1000 test samples. \textbf{Tiny Imagenet}~\cite{wu2017tiny} datasets contains ten tasks, where each task is now a twenty-way classification. It is a subset of 200 classes from Imagenet~\cite{deng2009imagenet}, where each image size is rescaled to 64×64. For each task, there are 8000 training samples, 2000 valid samples and 1000 test samples. For continual semantic segmentation: \textbf{Pascal VOC 2012} ~\cite{vocdataset} is consisted of 20 object classes plus the background class. Specifically, we resize the images to 512×512 with a random resize crop and apply an additional random horizontal flip augmentation at the training time. Moreover, the images are resized to 512×512 with a center crop when testing the segmentation ability of the model.

\subsubsection{Implementation Details} The experiments are implemented in Pytorch and conducted on one NVIDIA GeForce RTX 3090 GPU. For a fair comparison, the same neural network architecture is used for all competing algorithms. For continual classification: about Permuted-MNIST dataset~\cite{EWC}, this is a multi-layer perception with 2 hidden layers of 400 nodes each and RELU non-linearities are used in all hidden layers. About CIFAR-100~\cite{krizhevsky2009learning} and Tiny Imagenet~\cite{wu2017tiny} datasets, SMALL model~\cite{delange2021continual} is used for all comparison algorithms. Based on a VGG configuration~\cite{chung2018implementation}, the SMALL model is with less parameters. SGD optimizer are adopted with $0.9$ momentum. We repeat each experiment ten times with different random seeds and give the average results. What is more, we adopt a grid search over all hyperparameters (\emph{e.g.,} search the learning rate in $[1e-2, 5e-3, 1e-3, 5e-4, 1e-4]$) to obtain best hyperparameters for each comparison algorithm. For continual semantic segmentation: following ~\cite{MIB,PLOP}, we use Deeplab-v3~\cite{deeplab} with a ResNet-101 backbone~\cite{resnet} pre-trained on ImageNet~\cite{imagenet} for all experiments. Each task contains 30 epochs for Pascal VOC 2012 with a batch size of 24. The learning rate of the first task in all experiments is $1e-2$, and that of the following tasks are $1e-3$. We reduce the learning rate exponentially with a decay rate of $9e-1$. We adopt SGD optimizer with $9e-1$ Nesterov momentum. The first task $t = 1$ is common to all algorithms for each setting, thus we reuse the weights trained in this task. In the phase of inference, no task id is provided, which is more realistic than some continual learning methods. 

\subsection{Experimental Results}

\begin{table*}[htbp]
\centering
\setlength{\tabcolsep}{1.0mm}
\caption{mIoU (\%) on Pascal-VOC 2012 under 10-1 overlapped scenario. \textcolor{purple}{\textbf{Red}}: the highest results.}
\resizebox{\linewidth}{!}{
	\begin{tabular}{c|ccccccccccccccccccccc|>{\columncolor{lightgray}}c}
		\toprule
		Class ID &0& 1 & 2 & 3 & 4 & 5 &6 & 7 & 8 & 9 & 10 & 11 & 12 & 13 & 14 & 15 & 16 & 17 & 18 & 19 & 20 & mIoU  \\
		\midrule
  
		Finetune & 68.9 & \, 0.0 & \, 0.0 & \, 0.0 & \, 0.0 & \, 0.0 & \, 0.0 & \, 0.0 & \, 0.0 & \, 0.0 & \, 0.0 & \, 0.0 & \, 0.0 & \, 0.0 & \, 0.0 & \, 0.0 & \, 0.0 & \, 0.0 & \, 0.0 & \, 0.0 & 10.4 & \, 3.8 \\

		EWC~\cite{EWC} & 72.9 & \, 0.0 & \, 0.0 & \, 0.0 & \, 0.0 & \, 0.0 & \, 0.0 & \, 0.0 & \, 0.0 & \, 0.0 & \, 0.0 & \, 0.0 & \, 0.0 & \, 0.0 & \, 0.0 & 48.7 & \, 0.0 & \, 0.0 & \, 0.0 & \, 0.0 & \, 0.0 & \, 5.8 \\

		LWF~\cite{LWF} & 79.3 & \, 0.0 & \, 0.0 & \, 0.0 & \, 0.0 & \, 0.0 & \, 0.0 & \, 0.0 & \, 0.0  & \, 0.0 & \, 0.0 & \, 0.0 & \, 0.0 & \, 0.0 & \, 0.0 & \, 9.6 & \, 0.0 & \, 0.0 & \, 7.9 & \, 8.4 & \, 7.1 & \, 5.4 \\

		LWF-MC~\cite{icarl} & 70.0 & \, 0.0 & \, 0.0 & \, 0.0 & \, 0.0 & \, 0.0 & \, 0.0 & \, 0.0 & \, 0.0 & \, 0.0 & \, 0.0 & \, 0.0 & \, 0.0 & \, 0.0 & \, 0.0 & \, 0.0 & \, 0.0 & \, 0.0 & \, 0.0 & 10.9 & 14.7 & \, 4.5 \\
  
		ILT~\cite{ILT} & 78.6 & \, 0.0 & \, 0.0 & \, 0.0 & \, 0.0 &  \, 0.0 & \, 0.0 & \, 0.0 & \, 0.0 & \, 0.0 & \, 0.0 & \, 0.0 & \, 0.0 & \, 0.0 & \, 0.0 & \, 5.0 & \, 0.0 & \, 0.0 & \, 8.4 & \, 7.8 & \, 8.6 & \, 5.2 \\

		MiB~\cite{MIB}  &82.1  & \, 0.0  & \, 7.2 & \, 0.6 & \, 0.0 & 23.3 & \, 0.1 & 13.8 & \, 2.8 & \, 0.1 & \, 0.0 & \, 0.0 & \, 0.0 & \, 0.1 & 12.8 & 76.5 & \, 0.0 & \, 8.8 & 13.7 & 13.1 & 18.5 & 13.0 \\
  
		PLOP~\cite{PLOP} & 67.5 & 60.0 & \, 0.7 & 16.4 & 30.2 & 48.7 & \, 2.4 & 71.7 & 50.1 & 15.9 & \, 0.0 & \, 4.2 & \, 0.1 & \, 3.3 & 27.1 & 67.7 & \, 9.4 & 11.4 & \, 7.0 & \, 9.5 & \, 2.6 & 24.1 \\

          ALIFE~\cite{alife} & 73.7  & \, 0.0 & \, 0.0 & \, 0.0 & \, 0.0 & \, 0.0 & \, 0.0 & \, 0.0 & \, 0.0 & \, 0.0 & \, 0.0 & \, 0.0 & \, 0.0 & \, 0.0 & \, 0.0 & \, 0.5 & \, 0.0 & \, 2.0 & \, 2.4 & 16.3 & 34.2 & \, 6.1 \\

		RCIL~\cite{RCIL} & 61.6 & 75.0 & 32.2 & 42.8 & 44.9 & 70.4 & 72.0 & 85.3 & 56.8 & 11.3 & \, 0.0 & 16.7 & 23.4 & \, 0.0 & 27.4 & 65.2 & \, 6.4 & \, 1.9 & \, 7.6 & 11.8 & \, 3.5 & 34.1 \\

		\midrule
		\textbf{RCIL~\cite{RCIL}+spWC} (Ours) & 86.0 & 80.6 & 39.0 & 51.9 & 48.2 & 64.3 & 48.3 & 85.8 & 66.0 & 17.4 & \, 0.0 & 24.5 & 33.0 & \, 0.8 & 41.4 & 76.5 & 20.6 & \, 8.8 & 20.9 & 23.4 & 16.3 & \textcolor{purple}{\textbf{40.7}} \\
		\bottomrule
\end{tabular}}
\label{tab: comparison_voc_10_1}
\end{table*}

\begin{table*}[htbp]
\centering
\setlength{\tabcolsep}{1.0mm}
\caption{mIoU (\%) on Pascal-VOC 2012 under 10-1 disjoint scenario. \textcolor{purple}{\textbf{Red}}: the highest results.}
\resizebox{\linewidth}{!}{
	\begin{tabular}{c|ccccccccccccccccccccc|>{\columncolor{lightgray}}c}
		\toprule
		Class ID &0& 1 & 2 & 3 & 4 & 5 &6 & 7 & 8 & 9 & 10 & 11 & 12 & 13 & 14 & 15 & 16 & 17 & 18 & 19 & 20 & mIoU  \\
		\midrule
  
		Finetune & 69.1 & \, 0.0 & \, 0.0 & \, 0.0 & \, 0.0 & \, 0.0 & \, 0.0 & \, 0.0 & \, 0.0  & \, 0.0 & \, 0.0 & \, 0.0 & \, 0.0 & \, 0.0 & \, 0.0 & \, 0.0 & \, 0.0 & \, 0.0 & \, 0.0 & \, 0.0 & \, 9.0 & \, 3.7\\

		EWC~\cite{EWC} & 73.9 & \, 0.0 & \, 0.0 & \, 0.0 & \, 0.0 & \, 0.0 & \, 0.0 & \, 0.0 & \, 0.0 & \, 0.0 & \, 0.0 & \, 0.0 & \, 0.0 & \, 0.0 & \, 0.0 & 42.1  & \, 0.0 & \, 0.0 & \, 0.0 & \, 0.0 & \, 0.0 & \, 5.5 \\

		LWF~\cite{LWF} & 77.0 & \, 0.0 & \, 0.0 & \, 0.0 & \, 0.0 & \, 0.0 & \, 0.0 & \, 0.0 & \, 0.0 & \, 0.0 & \, 0.0 & \, 0.0 & \, 0.0 & \, 0.0 & \, 0.0 & \, 7.0 & \, 0.0 & \, 0.0 & \, 8.2 & \, 6.4 & \, 6.8 & \, 5.0 \\

		WF-MC~\cite{icarl} & 75.1 & \, 0.0 & \, 0.0 & \, 0.0 & \, 0.0 & \, 0.0 & \, 0.0 & \, 0.0 & \, 0.0 &\, 0.0  & \, 0.0 & \, 0.0 & \, 0.0 & \, 0.0 & \, 0.0 & \, 0.1 & \, 0.0 & \, 0.2 & \, 0.0 & \, 6.6 & \, 4.9 & \, 4.1 \\
  
		ILT~\cite{ILT} & 81.7 & \, 0.0 & \, 0.0 & \, 0.0 & \, 0.0 & \, 0.0 & \, 0.0 & \, 0.0 & \, 0.0 & \, 0.0 & \, 0.0 & \, 0.0 & \, 0.0 & \, 0.0 & \, 0.0 & \, 2.6 & \, 0.0 & \, 0.2 & \, 7.0 & \, 6.6 & \, 9.1 & \, 5.1 \\

		MiB~\cite{MIB}  & 69.0 & \, 0.0 & \, 8.6 & \, 0.0 & \, 0.2 & 17.7 & \, 0.0 & \, 0.7 & \, 0.2 & \, 0.2 & \, 0.0 & \, 4.6 & \, 0.0 & \, 0.0 & \, 6.5 & 46.7 & \, 0.0 & \, 9.1 & \, 5.8 & \, 9.0 & \, 6.8 & \, 8.8 \\
  
		PLOP~\cite{PLOP} & \, 0.0 & \, 1.5 & 19.0 & \, 4.4 & 18.2 & 34.5 & \, 1.8 & 62.7 & 27.8 & 13.2 & \, 0.0 & \, 5.2 & \, 0.1 & \, 0.0 & 11.9 & 66.7 & \, 0.0 & \, 4.9 & \, 2.2 &  11.4 & \, 3.6 & 13.8 \\

		ALIFE~\cite{alife} & 75.3 & \, 0.0 & \, 0.0  & \, 0.0 & \, 0.0 & \, 0.0 & \, 0.0 & \, 0.0 & \, 0.0 & \, 0.0 & \, 0.0 & \, 0.0 & \, 0.0 & \, 0.0 & \, 0.3 & \, 0.0 & \, 0.0 & \, 0.1 & 14.4 & 13.9 & 27.8 & \, 5.7 \\
  
		RCIL~\cite{RCIL} & 43.0 & 59.9 & 16.6 & 63.8 & 26.6 & 41.8 & 17.6 & 56.0 & \, 9.6 & \, 5.6 & 14.8 & \, 2.3 & \, 2.8 & \, 0.3 & \, 7.8 & \, 6.0 & \, 1.3 & 12.8 & \, 4.2 & 18.7 & \, 6.1 & 20.2\\

		\midrule
		\textbf{RCIL~\cite{RCIL}+spWC} (Ours) & 81.3 & 54.4 & 21.2 & 47.3 & 37.2 & 52.0 & 12.9 & 54.1 & \, 8.8 & 10.4 & 25.1 & \, 3.2 & \, 3.7 & \, 0.0 & 11.3 & 23.8 & \, 5.5 & \, 16.0 & 14.5 & 23.5 & 12.7 & \textcolor{purple}{\textbf{24.2}}\\
		\bottomrule
\end{tabular}}
\label{tab: comparison_voc_10_1_disjoint}
\end{table*}

This subsection presents extensive experimental results on four popular datasets, \emph{i.e.,} Permuted-MNIST~\cite{EWC}, CIFAR-100~\cite{krizhevsky2009learning}, Tiny Imagenet~\cite{wu2017tiny} and Pascal VOC 2012~\cite{vocdataset}.
\begin{figure*}[ht]
\centering
\hspace{-4mm}
\subfigure[Image]{
\label{fig:subfig:1}
\includegraphics[scale=0.71]{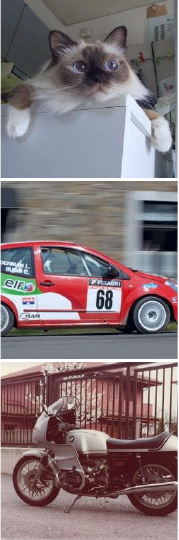}}
\hspace{-4mm}
\subfigure[ILT~\cite{ILT}]{
\label{fig:subfig:2}
\includegraphics[scale=0.71]{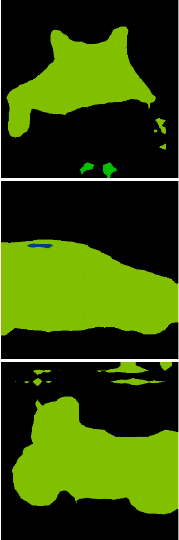}}
\hspace{-4mm}
\subfigure[MiB~\cite{MIB}]{
\label{fig:subfig:3}
\includegraphics[scale=0.71]{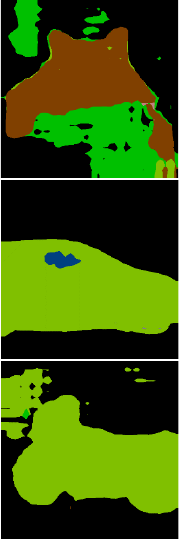}}
\hspace{-4mm}
\subfigure[PLOP~\cite{PLOP}]{
\label{fig:subfig:4}
\includegraphics[scale=0.71]{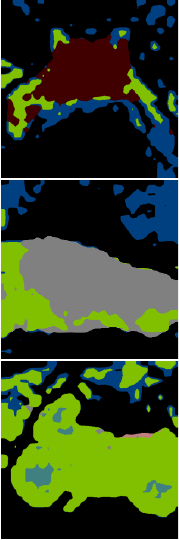}}
\hspace{-4mm}
\subfigure[ALIFE~\cite{alife}]{
\label{fig:subfig:5}
\includegraphics[scale=0.71]{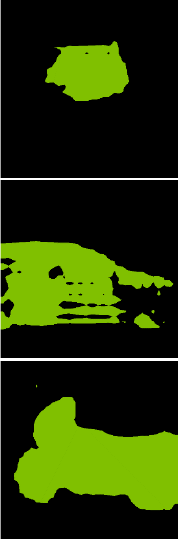}}
\hspace{-4mm}
\subfigure[RCIL~\cite{RCIL}]{
\label{fig:subfig:6}
\includegraphics[scale=0.71]{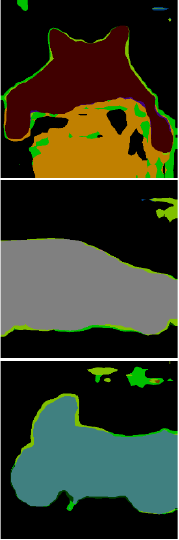}}
\hspace{-4mm}
\subfigure[RCIL+spWC]{
\label{fig:subfig:7}
\includegraphics[scale=0.71]{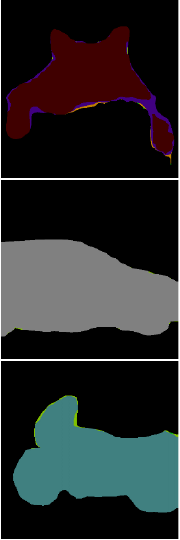}}
\hspace{-4mm}
\subfigure[GT]{
\label{fig:subfig:8}
\includegraphics[scale=0.71]{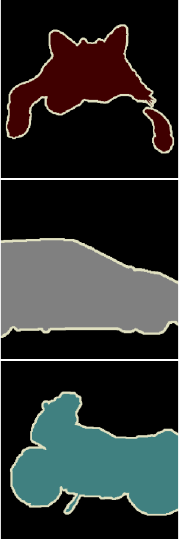}}

\caption{The qualitative comparison between state-of-the-art algorithms. All results are from the last step of the 10-1 overlapped scenario on the Pascal VOC 2012 dataset. The visualization results demonstrate the effectiveness of our spWC framework.}

\label{fig:vis}
\end{figure*}

\subsubsection{Accuracy Evaluation} We compare all the comparison algorithms with our proposed EWC+spWC and MAS+spWC on Permuted-MNIST~\cite{EWC}, CIFAR-100~\cite{krizhevsky2009learning} and Tiny Imagenet~\cite{wu2017tiny} datasets to show the classification capabilities.

From the presented Average-\textit{APA} in ~\cref{table_accuracy}, we have the following observations. 1) Except for JointTrain, our proposed EWC+spWC or MAS+spWC can achieve the best performance on all three datasets, which can obtain more than 1.2\% improvement on Permuted-MNIST, 1.5\% improvement on CIFAR-100 and 2.3\% improvement on Tiny Imagenet dataset, respectively. This observation justifies the effectiveness of consolidating previous tasks discriminatively. Although JointTrain is the best performer in terms of Average-\textit{APA}, saving all training samples and training from scratch needs abundant storage space and training time, and also faces the problem of data privacy. Given the inevitable shortcomings of JointTrain, we consider it to be not an optimal algorithm and no longer compare with it. 2) Compared with EWC, our proposed EWC+spWC achieves 2.1\% improvement on Permuted-MNIST, 4.0\% improvement on CIFAR-100 and 3.2\% improvement on Tiny Imagenet dataset. Compared with MAS, our proposed spMS achieves 3.7\% improvement on Permuted-MNIST, 4.8\% improvement on CIFAR-100 and 5.4\% improvement on Tiny Imagenet dataset. The main reason is that our EWC+spWC and MAS+spWC can distinguish the latent relationship between previous tasks and the new task, and select ``difficult" previous tasks to consolidate in the training phase for the new task. 3) Compared with online-EWC, our proposed spWC achieves 1.6\% improvement on Permuted-MNIST, 4.8\% improvement on CIFAR-100 and 3.8\% improvement on Tiny Imagenet dataset. It is mainly caused by the essence of online-EWC itself, \emph{i.e.,} the contribution of the calculated parameter importance of the previous task to the new task gradually decreases, which leads to serious forgetting of online-EWC. 4) Additionally, our EWC+spWC and MAS+spWC are both robust continual learners, since they perform stably on three datasets with different levels of difficulty (\emph{i.e.,} simple Permuted-MNIST dataset, difficult CIFAR-100 dataset and more difficult dataset Tiny Imagenet). Overall, without sacrificing data privacy and training time, the accuracy of our proposed EWC+spWC and MAS+spWC algorithms surpasses other state-of-the-art algorithms, which illustrates the effectiveness of our spWC framework.

\begin{figure}[t]
	\centering
	
	\subfigure[{\textit{Parameters Size (PS)} efficiency of spWC on Permuted-MNIST dataset}]
	{
		\label{a_1} 
		\includegraphics[width=0.48\textwidth]{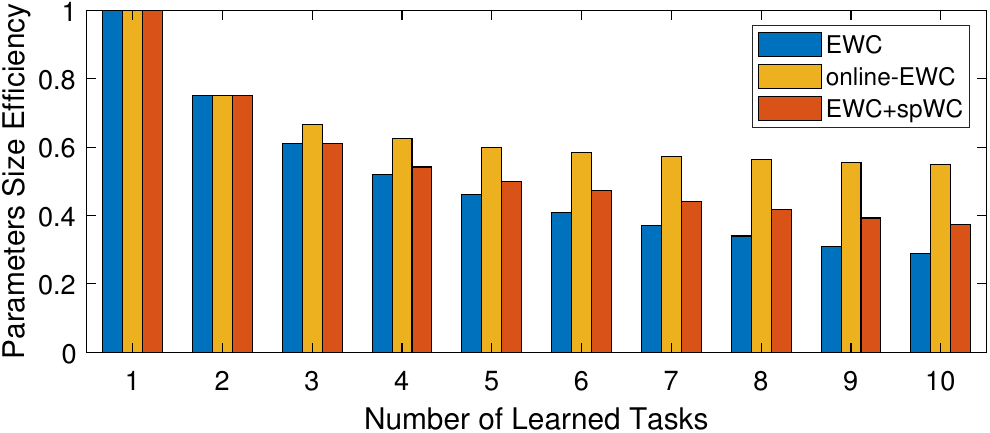}
	}
	
	\subfigure[{\textit{Parameters Size (PS)} efficiency of spMS on Permuted-MNIST dataset}]
	{
		\label{a_2} 
		\includegraphics[width=0.48\textwidth]{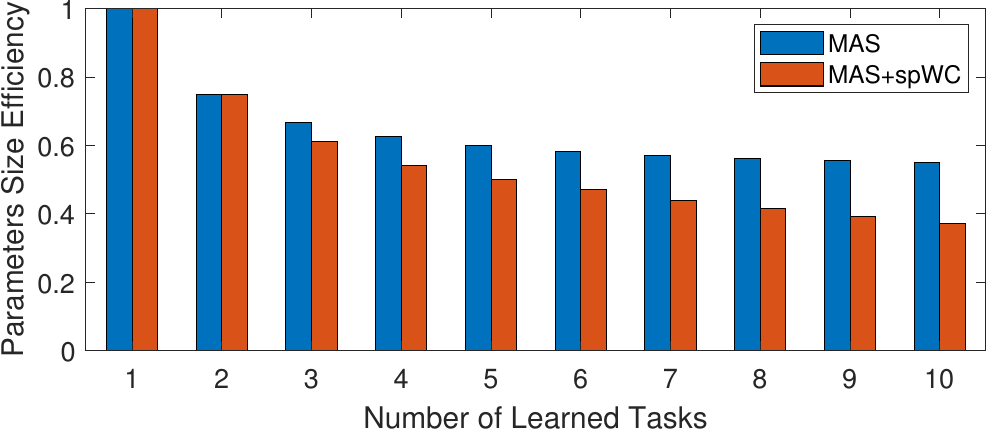}
	}
	
	\caption{{Comparison results of our proposed EWC+spWC (\textbf{Top}) and MAS+spWC (\textbf{Bottom}) when training each task on the Permuted-MNIST with the \textit{Parameters Size (PS)} efficiency.}}
	\label{permuted_e} 
\end{figure}

From the presented Average-\textit{ACF} in ~\cref{table_accuracy}, we have the following observations. 1) EWC+spWC and MAS+spWC possess the strongest anti-forgetting ability, since self-paced learning makes our EWC+spWC and MAS+spWC continuously store knowledge via imitating ``human learning'', \emph{i.e.,} learning from simple to difficult. 2) Compared with EWC, our proposed EWC+spWC has increased anti-forgetting ability by 2.3\% on Permuted-MNIST, 4.4\% on CIFAR-100 and 3.5\% on Tiny Imagenet dataset. Compared with MAS, our proposed spMS has increased anti-forgetting ability by 4.1\% on Permuted-MNIST, 5.2\% on CIFAR-100 and 5.9\% on Tiny Imagenet dataset. This is because that properly selecting ``difficult" previous tasks to consolidate emphatically in the training phase can prevent interference between tasks and improve the memory ability of the model. 3) Compared with online-EWC, our proposed EWC+spWC has increased anti-forgetting ability by 1.7\% on Permuted-MNIST, 5.3\% on CIFAR-100 and 4.2\% on Tiny Imagenet dataset. This observation verifies the superiority of EWC+spWC. 4) Moreover, we can also observe from ~\cref{table_accuracy} that the performance of our EWC+spWC and MAS+spWC algorithms is better than other comparison algorithms in terms of Average-\textit{ACF}, which can well illustrate the generalization of our proposed spWC framework. In general, our proposed EWC+spWC and MAS+spWC are the most comprehensive algorithms among all comparison algorithms.

From the presented \textit{APA} and \textit{ACF} in ~\cref{he}, we have the following observations. 1) As the number of tasks increases, except for JointTrain, the performance of all algorithms has declined, because JointTrain can access samples of all tasks. However, other algorithms have problems of forgetting since they only rely on training data of the current task. 2) The metric \textit{APA} of the proposed spWC and spMS algorithms has only a small decrease, while the performance of other competing algorithms has a relatively large decrease. This phenomenon is due to the discrepancy in data distribution between new task and previous tasks, and treating the contribution of continuous tasks equally will degrade the model performance. 3) As the number of tasks increases, the metric \textit{ACF} of our EWC+spWC and MAS+spWC has always remained at the best except for JointTrain, which further shows that the proposed spWC framework are less subject to external interference.

\begin{figure}[t]
	\centering
	\includegraphics[width=1\columnwidth]{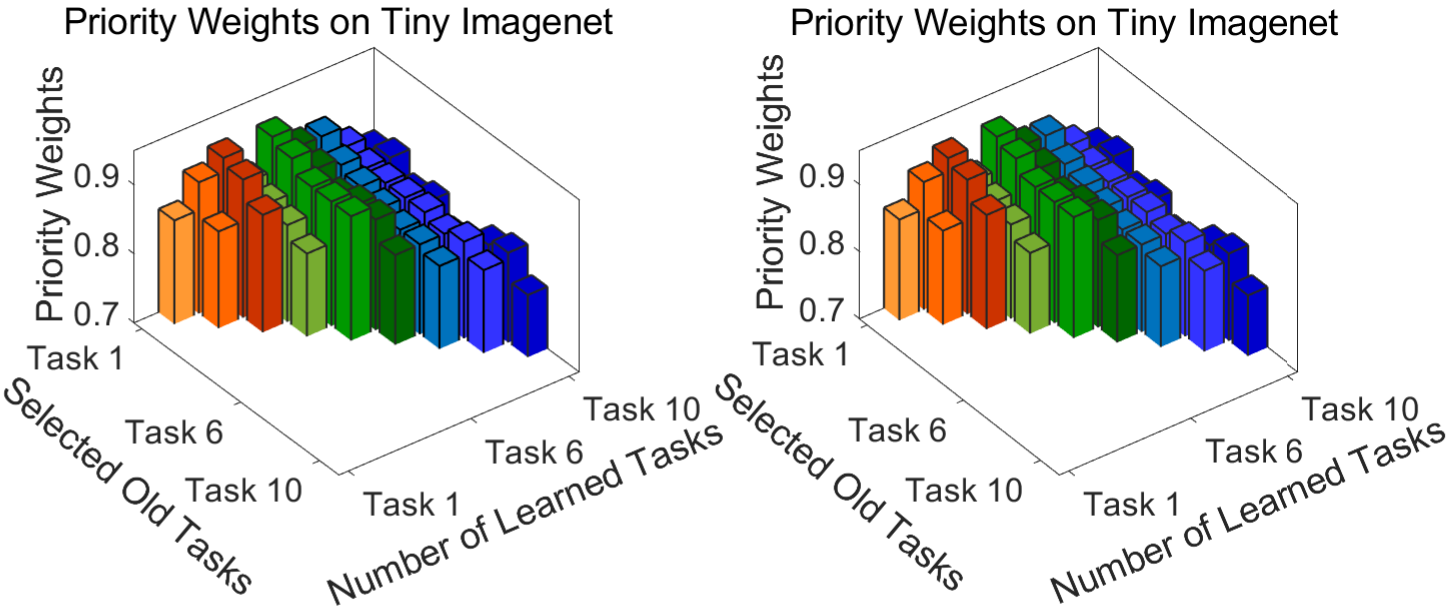} 
	
	\caption{{Values of priority on the Tiny Imagenet obtained by our proposed EWC+spWC (\textbf{Left}) and MAS+spWC (\textbf{Right}). }}
	\label{tiny_weight} 
\end{figure}

\begin{figure}[t]
	\centering
	\includegraphics[width=1\columnwidth]{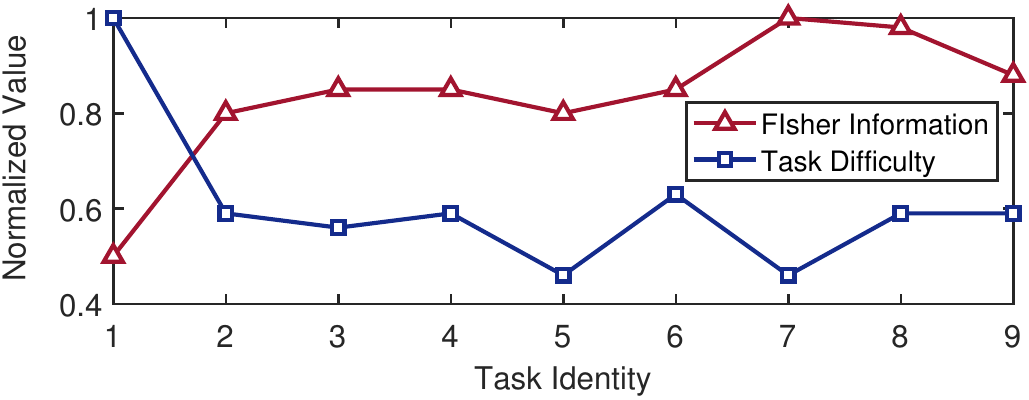} 
	
	\caption{{Illustration of the relation between FIsher information and task difficulty on the Permuted-MNIST dataset.}}
	\label{relation}
\end{figure}

From the presented mIoU in ~\cref{tab: comparison_voc_10_1} and ~\cref{tab: comparison_voc_10_1_disjoint}, our RCIL+spWC excels in different continual semantic segmentation scenarios, \emph{i.e.,} \textit{disjoint} and \textit{overlapped} settings. If the pixels in the background class at step $t$ belong to old classes and the true background class, it is a \textit{disjoint} setting. If the pixels in the background class belong to old classes, future classes and the true background class, it is an \textit{overlapped} setting. The testing classes are labeled for all seen classes to evaluate the current model's incremental learning performance. On Pascal VOC 2012 dataset~\cite{vocdataset},  10-1 scenario denotes 10 classes followed by 1 class ten times. Specifically, our RCIL+spWC outperforms the state-of-the-art algorithms by 6.6\% and 4.0\% in terms of mIoU in the 10-1 \textit{overlapped} and 10-1 \textit{disjoint} scenarios, respectively. Besides, we present the visualization results of competing algorithms in the 10-1 overlapped scenario in ~\cref{fig:vis}. Our segmentation results are much better than other algorithms, illustrating that the proposed spWC framework are robust on other continual learning directions, \emph{e.g.,} continual semantic segmentation.

\subsubsection{Efficiency Evaluation} In ~\cref{permuted_e}, we compare the metric \textit{PS} which can reflect computational efficiency of different algorithms. We have the following observations. 1) When the number of tasks is 1, the computational efficiency of EWC+spWC and MAS+spWC is equal to that of EWC and MAS respectively, since the algorithm for training the first task is the same. 2) The computational efficiency of EWC+spWC is always higher than that of EWC, which shows the efficiency of our proposed spWC framework. 3) As the number of tasks increases, the computational efficiency of EWC+spWC is improved compared with EWC (\emph{e.g.,} when the task number is 10, the computation efficiency of EWC+spWC improves from 0.29 to 0.46 compared with EWC) and the advantage of our proposed EWC+spWC becomes more and more obvious. This is because EWC+spWC only needs parameters of selected previous tasks to participate in learning the new task, while EWC needs the parameters of all previous tasks. 4) The efficiency of EWC+spWC and MAS+spWC lag behind online-EWC and MAS, which comes at the cost of sacrificing accuracy for online-EWC and MAS. Therefore, our spWC framework is a good balance between accuracy and efficiency.

\subsubsection{Values of Priority} In order to better express that our proposed EWC+spWC and MAS+spWC can reasonably allocate the priority weight to each past task, we provide the values of priority on Tiny Imagenet dataset in ~\cref{tiny_weight}, and have the following observations. 1) As can be seen from ~\cref{tiny_weight}, the priority weight of each old task will change as the number of learned tasks increases. This is because that self-paced learning can adaptively adjust the task weight during the learning process. 2) As the number of tasks increases, the old task priority weight of EWC+spWC and MAS+spWC tends to increase because of the more forgetting of old tasks. 3) The priority weight distribution of our proposed algorithms is approximate on account of the identical task difficulty.

\subsubsection{Grad-CAM Visualization on Task Difficulty} To determine which tasks are ``easy'' and ``difficult'', we have conducted experiments with Grad-CAM visualization (\emph{i.e.,} the heatmap of the activation class generated for the input image). The results are given in \cref{fig:cam}. Specifically, the Grad-CAM visualization can be understood as the contribution distribution to the predicted output of the model. A higher score of the corresponding area in the input image (\emph{i.e.,} area that is darker in color) represents a greater contribution to the classification model. Intuitively, the face and eyes (\emph{i.e.,} the unique and ubiquitous features) of the task \textit{cat} are visually distinct, making it relatively easy to be learned and captured by the model. However, the screen features of the task \textit{monitor} are difficult to be captured, and the main reason is that they are relatively abstract and diverse. Moreover, the task \textit{table} is also difficult to learn, since the model is too dependent on the features of the food on it, while ignoring to dig out the features of \textit{table} itself. Therefore, experimental results of Grad-CAM visualization demonstrate that the tasks encountered in continual learning are in varying degrees of difficulty due to diverse feature representations for different classes by the model.

{\subsubsection{Relation between FIsher Information and Task Difficulty}In order to make our proposed spWC framework more convincing, we visualize the relation between ``task difficulty" and ``value of FIsher information" in ~\cref{relation}, where the values of them are normalized. Specifically, the value of ``task difficulty" refers to the difficulty of the nine tasks relative to the $10$-th task; the value of ``FIsher information" is obtained by summing the FIsher information of the task, and then normalizing the value to [0,1]. Obviously, we can observe that FIsher information is not correlated to task difficulty. Therefore, the original algorithms (\emph{e.g.,} EWC and MAS) cannot treat tasks differently according to task difficulty, which indicates the importance of our spWC framework.}

\begin{figure}[t]
\centering
\hspace{-3mm}
\subfigure[Task 1: cat]{
\label{fig:subfig:a}
\includegraphics[scale=0.91]{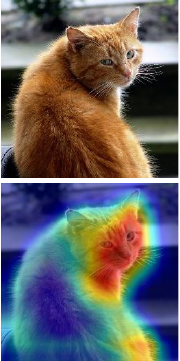}}
\hspace{-2mm}
\subfigure[Task 2: monitor]{
\label{fig:subfig:c}
\includegraphics[scale=0.91]{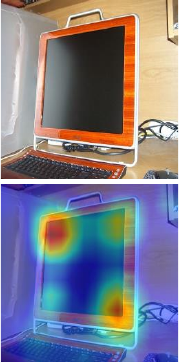}}
\hspace{-2mm}
\subfigure[Task 3: table]{
\label{fig:subfig:d}
\includegraphics[scale=0.91]{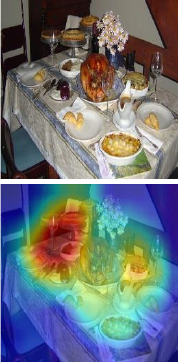}}
\caption{Grad-CAM visualization to determine which tasks are ``easy'' and ``difficult''. Area in dark color means high score.} 
\label{fig:cam}
\end{figure}

\begin{table}[t]
	\caption{{ Self-paced Regularizations and the Corresponding Close-formed Priority Weights.}}
	\scalebox{0.9}{
		\begin{tabular}{c|c|c}
			\toprule
			Regularizations & Self-paced Regularizations & Priority Weights \\ \midrule
			Hard~\cite{kumar2010self}                  & $-\lambda\sum_{i=1}^{m-1} v_i$ & $
			\left\{
			\begin{array}{lr}
				1, \quad l_t<\mu_m\\
				
				0,\quad {\rm otherwise}
			\end{array}
			\right.$ \\ \midrule
			Linear~\cite{jiang2014easy}                & $\frac{1}{2}\lambda\sum_{i=1}^{m-1} (v_i^2-2v_i)$ & $
			\left\{
			\begin{array}{lr}
				1-l_t/\lambda, \quad l_t<\mu_m\\
				
				0,\quad {\rm otherwise}
			\end{array}
			\right.$ \\ \midrule
			Logarithmic~\cite{jiang2014easy}           & \tabincell{l}{$\sum_{i=1}^{m-1} (\zeta v_i-\frac{\zeta v_i}{log\zeta})$ \\ $\zeta=1-\mu_m, 0<\mu_m<1$} & $
			\left\{
			\begin{array}{lr}
				\frac{log(l_t+\zeta)}{log\zeta}, \quad l_t<\mu_m\\
				
				0,\quad {\rm otherwise}
			\end{array}
			\right.$ \\ \bottomrule
		\end{tabular}
	}
    
    \label{table_regularizers}
\end{table}

\begin{figure}[t]
	\centering
	
	\subfigure[{\textit{Average Per-Task Accuracy} of EWC+spWC on Tiny Imagenet dataset}]
	{
		\label{f_1} 
		\includegraphics[width=0.48\textwidth]{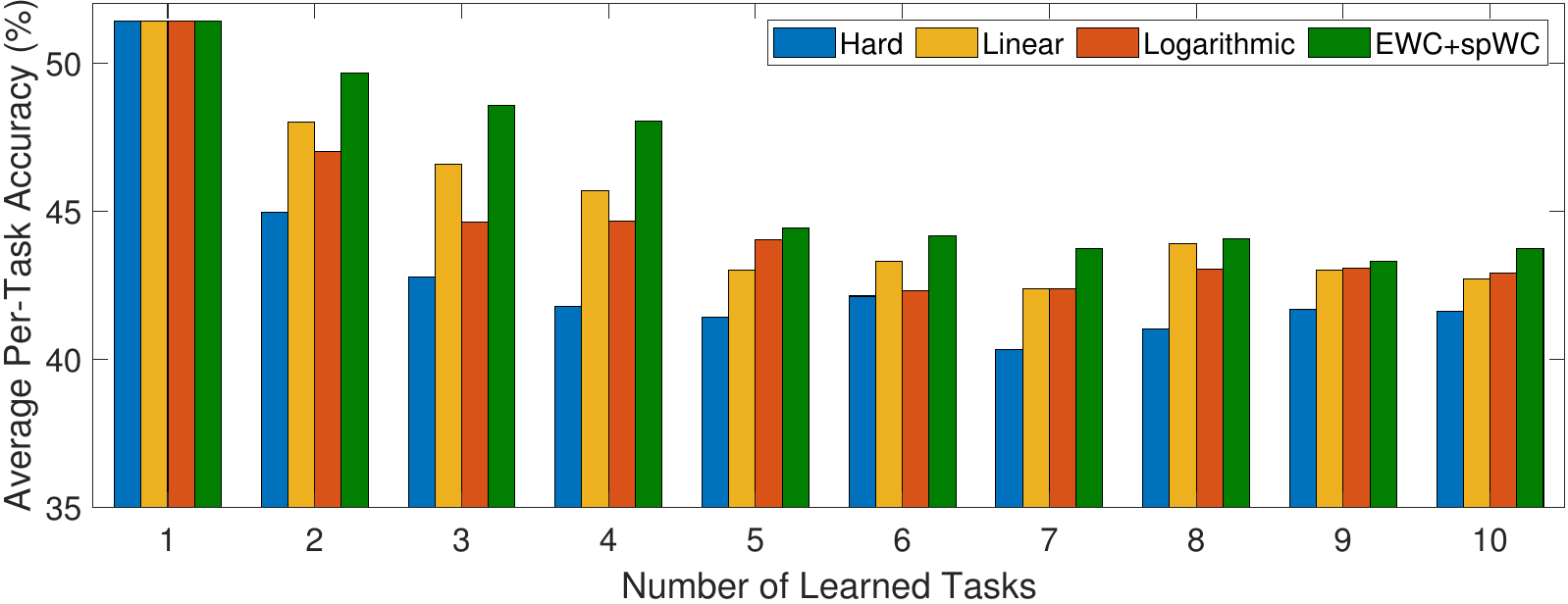}
	}
	
	\subfigure[{\textit{Average Per-Task Accuracy} of MAS+spWC on Tiny Imagenet dataset}]
	{
		\label{f_2} 
		\includegraphics[width=0.48\textwidth]{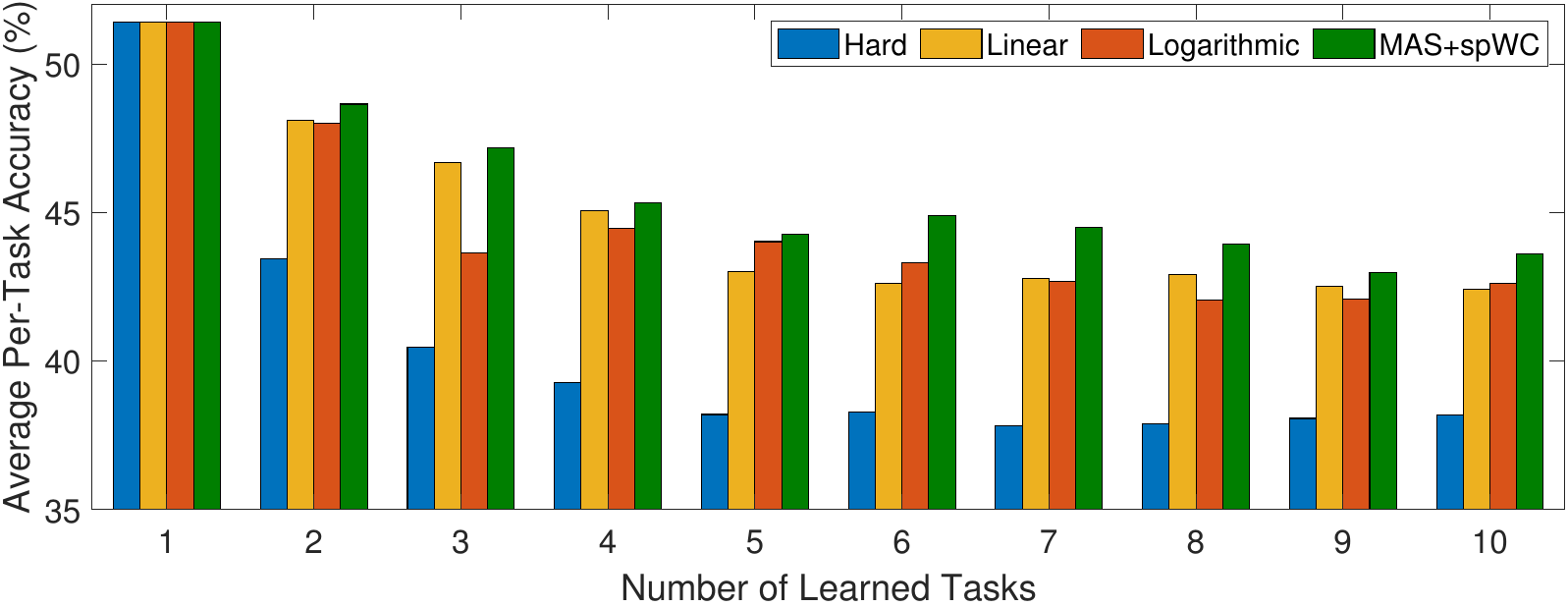}
	}
	
	\caption{{\textit{Average Per-Task Accuracy} for different self-paced regularizations of our proposed EWC+spWC (\textbf{Top}) and MAS+spWC (\textbf{Bottom}) on the Tiny Imagenet dataset. }}
	\label{tiny_f} 
\end{figure}

\begin{figure*}[htbp]
	\centering
	\includegraphics[width=2.06\columnwidth]{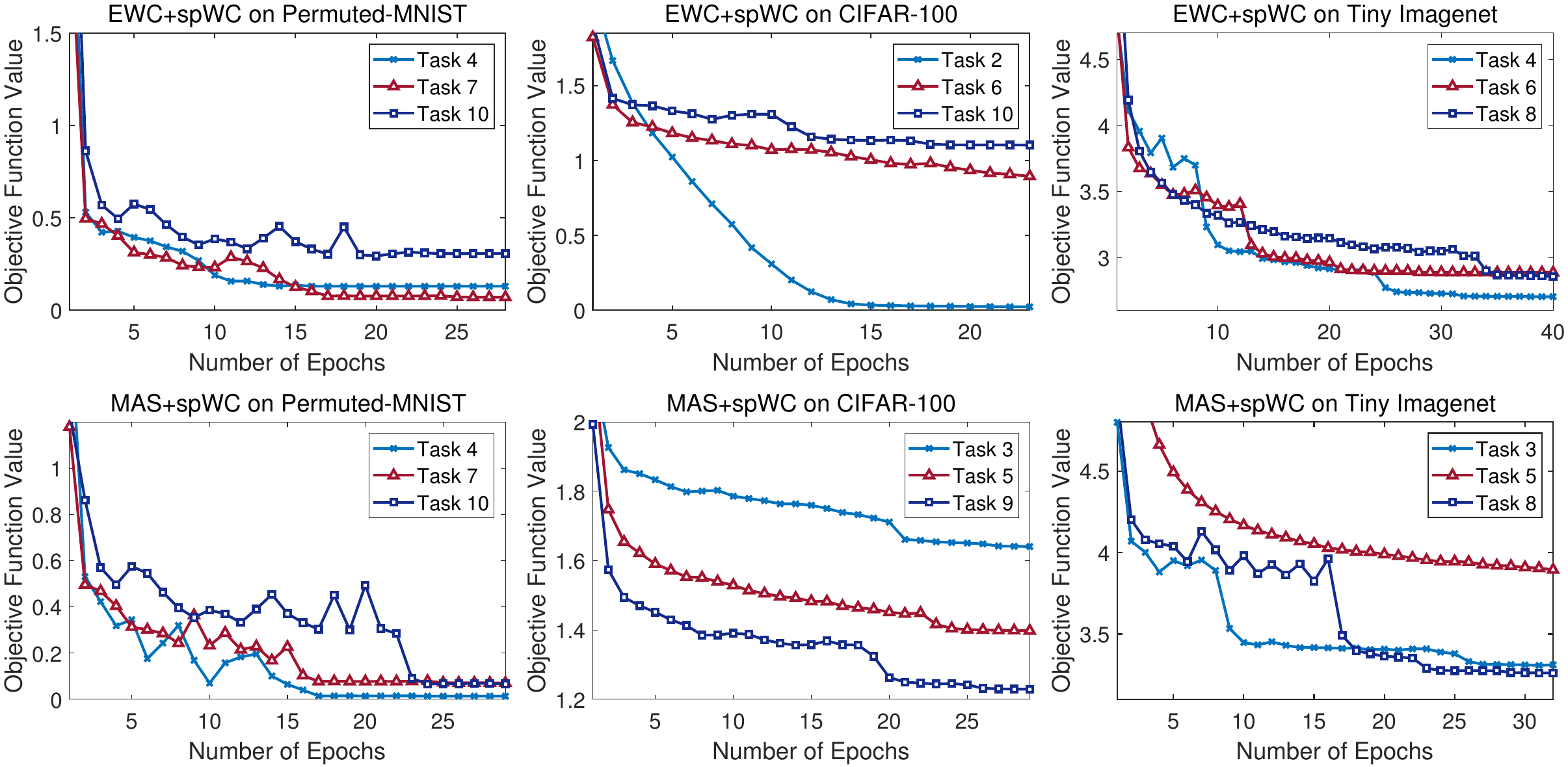} 
	
	\caption{{Convergence analysis of our proposed EWC+spWC (\textbf{Top}) and MAS+spWC (\textbf{Bottom}) on three public datasets, \emph{i.e.,} Permuted-MNIST (\textbf{Left}), CIFAR-100 (\textbf{Middle}) and Tiny Imagenet (\textbf{Right}), where different colors denote different tasks.}}
	\label{convergence}
	
\end{figure*}

\subsection{Ablation Study} 
\subsubsection{Self-paced Regularization}In this subsection, we compare the proposed self-paced regularization with the hard weighting~\cite{kumar2010self}, linear soft weighting and logarithmic soft weighting~\cite{jiang2014easy}. {The specific definitions of self-paced regularizations and the corresponding close-formed priority weights are provided in ~\cref{table_regularizers}. 
The \textit{Average Per-Task Accuracy (APA)} after training each task based on different self-paced regularizations are provided in ~\cref{tiny_f}. We have the following observations. 1) The \textit{Average Per-Task Accuracy (APA)} with the proposed self-paced regularization achieves 0.8\% and 1.0\% improvement on Tiny Imagenet dataset compared with Logarithmic, respectively obtained by our EWC+spWC and MAS+spWC. This confirms the rationality of the priority weight given to each task by our proposed self-paced regularization. 2) Hard weighting is the worst performance of all regularizations, because hard weighting does not adjust task weights due to different task difficulty, which also illustrates the importance of adaptively adjusting weights. 3) As the number of tasks increases, the advantages of our self-paced regularization are always maintained, {which confirms that our proposed self-paced regularization is best.}

\begin{table}[ht]
\caption{The mIoU(\%) of the last step on Pascal VOC 2012 for 10-1 overlapped scenario by selecting $k$ previous tasks.}
\setlength\tabcolsep{12pt}
\centering
\small
\scalebox{0.915}{
\begin{tabular}{ccc|ccc}
 \hline
\multicolumn{6}{c}{\textbf{10-1 Overlapped}}      \\ \hline
\multicolumn{3}{c|}{\textbf{$k=5$}} & \multicolumn{3}{c}{\textbf{$k=10$}}  \\ \hline 
0-10        & 11-20       & all       & 0-10        & 11-20        & all    \\ \hline
43.5        & 20.2        & 32.4      & 50.1        &  22.9        &  37.1    \\ 
 \hline \hline
\multicolumn{3}{c|}{\textbf{$k=15$}} & \multicolumn{3}{c}{\textbf{$k=20$}}  \\ \hline 
0-10        & 11-20       & all       & 0-10        & 11-20        & all   \\ \hline
53.4        & 26.6        & 40.7      & 53.4        &  26.3        &  40.5 \\ \hline
\end{tabular}}
\label{tab:k}
\end{table}

\subsubsection{Selection of $k$} The number of old tasks (\emph{i.e.,} $k$) chosen to participate in the computation of the regularization term in continual learning is important. We provide the comparison results by selecting different $k$ in~\cref{tab:k}. Experimental results show that our spWC framework achieves the best performance when $k=15$. In addition, when $k$ is large, the performance of our spWC framework is higher than state-of-the-art algorithms. This is because treating old tasks differently improves model performance. However, when the value of $k$ is small (\emph{{i.e.,}} $k=5$), the regularization term only constrains a small number of tasks, which causes the model to forget most tasks. Specifically, the computation complexity of our spWC is $O(\xi(6e7, 5e2 )+4e11)$ when $k=15$, which is efficient than $O(\xi(6e7, 5e2 )+6e11)$ of other algorithms, \emph{e.g.,} RCIL.

\subsection{Convergence Analysis}
In order to analyze the convergence of the objective function, we demonstrate the objective function value of EWC+spWC and MAS+spWC on Permuted-MNIST, CIFAR-100 and Tiny Imagenet datasets. As shown in ~\cref{convergence}, we have the following observations: 1) The objective function values for each new task decrease with respect to epoch number, where the values for each new task can quickly approach to a fixed point just using a few epoches (\emph{e.g.,} about 20 epochs for all tasks on all adopted datasets). 2) The more difficult the task, the higher the convergence value of the model. Therefore, experiments for tasks of different difficulty can better illustrate the robustness of our proposed EWC+spWC and MAS+spWC. 3) The objective function value can remain stable with little disturbance, which validates the convergence of the proposed optimization algorithm. In addition, the final objective function value and convergence epochs among different datasets vary lightly, since task distributions in each dataset are different. Consequently, ~\cref{convergence} can indirectly guarantee the convergence of our proposed EWC+spWC and MAS+spWC.

\section{Conclusion}
\label{conclusion}
This paper studies how to address the problems of equally treating previously learned tasks and large computational cost in continual learning. To be specific, we propose a self-paced Weight Consolidation (spWC) framework, which could evaluate the discriminative contributions of previous tasks to the new task for robust continual learning. The core spWC framework is applicable to all regularization-based continual learning algorithms. In the proposed EWC+spWC, MAS+spWC and RCIL+spWC, the difficulty of past learned tasks is measured by the key performance indicator (\emph {i.e.}, accuracy). Then each previous task is given a priority weight by the designed self-paced regularization to reflect its importance to the new task. Therefore, the parameters of new task will be learned via selectively consolidating the knowledge among past ``difficult" tasks. An alternative convex search is adopted to iteratively update the model parameters and priority weights in the bi-convex formulation. We have conducted experiments on Permuted-MNIST, CIFAR-100, Tiny Imagenet and Pascal VOC 2012 datasets. The experimental results demonstrate the effectiveness and efficiency of our spWC framework.

While promising, we admit our work has some limitations as follows. First, there lacks some theoretical analysis when applying the proposed spWC framework to continual learning algorithms for guaranteeing the convergence in the theory perspective. Second, although outperforming other state-of-the-art algorithms with a large margin, the proposed spWC framework has a poor performance in the continual learning process with long challenging tasks (\emph{e.g.}, continual semantic segmentation in~\cref{tab: comparison_voc_10_1} and \cref{tab: comparison_voc_10_1_disjoint}). In the future, we will focus on addressing these limitations, and extend our framework to 3D classification~\cite{9524506} and object detection~\cite{CDT}.
\ifCLASSOPTIONcaptionsoff
  \newpage
\fi



\bibliographystyle{IEEEtran}
\bibliography{self-paced-weight-consolidation-0117}
\end{document}